\documentclass[final]{cvpr}
\usepackage{times}
\usepackage{epsfig}
\usepackage{graphicx}
\usepackage{amsmath}
\usepackage{amssymb}
\usepackage{subfigure}
\usepackage{algpseudocode} 
\usepackage{multirow}

\newcommand*{\affaddr}[1]{#1} 
\newcommand*{\affmark}[1][*]{\textsuperscript{#1}}
\newcommand*{\email}[1]{\texttt{#1}}
\usepackage{caption}
\makeatletter
\newcommand{\printfnsymbol}[1]{%
  \textsuperscript{\@fnsymbol{#1}}%
}
\makeatother

\usepackage[pagebackref=true,breaklinks=true,colorlinks,bookmarks=false]{hyperref}

\begin{document}

\title{Skimming and Scanning for Untrimmed Video Action Recognition}




\author{%
Yunyan Hong\affmark[1]\thanks{Equal contribution}, Ailing Zeng\affmark[2]\printfnsymbol{1}, Min Li\affmark[2], Cewu Lu\affmark[1], Li Jiang\affmark[1], Qiang Xu\affmark[2]\\
\affaddr{\affmark[1]Shanghai Jiao Tong University}\\
\affaddr{\affmark[2]The Chinese University of Hong Kong}\\
\email{\tt\small\{hongyunyan, lucewu, ljiang\_cs\}@sjtu.edu.cn} \\ 
\email{\tt\small\{alzeng, mli, qxu\}@cse.cuhk.edu.hk}
}

\twocolumn[{%
\renewcommand\twocolumn[1][]{#1}
\maketitle
\vspace{-35pt}
\begin{center}
    \centering
    \captionsetup{type=figure}
    \includegraphics[width=\textwidth]{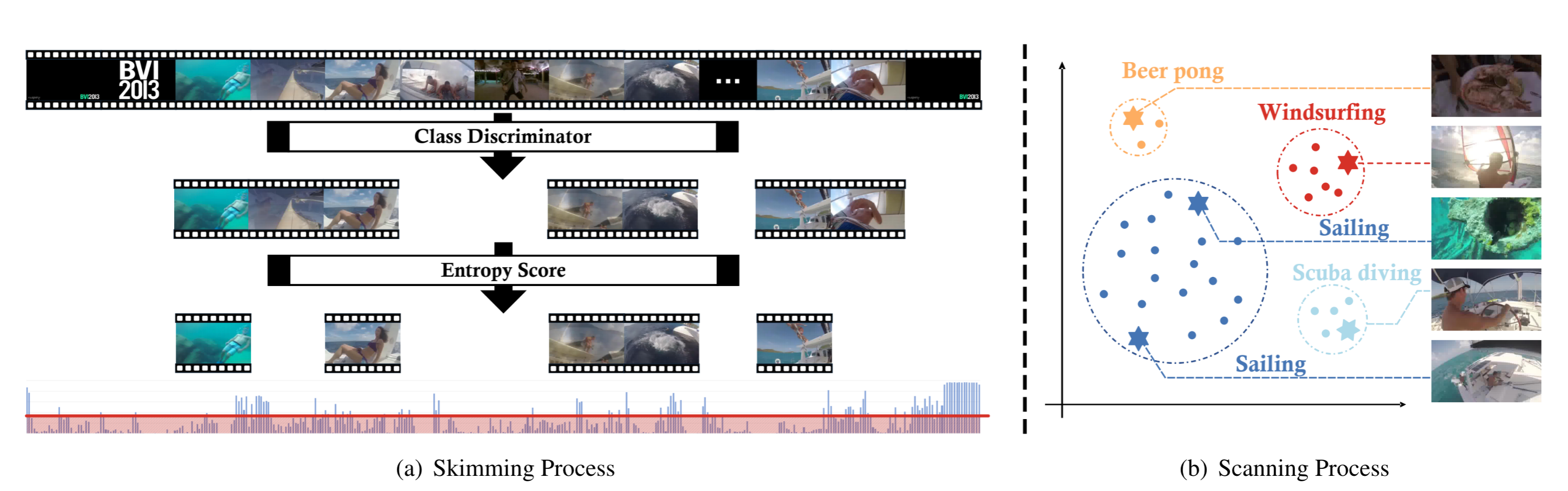}
    \caption{(a) shows the skimming process for a video of the “Sailing” action, (b) shows the scanning process for the remaining video clips after skimming. In (a), we eliminate those misleading clips and the clips whose entropy is higher than a pre-defined threshold (the red line),
    and the rest clips are represented as the points and stars in (b), in which the same color represents the same label. We select those diverse stars in scanning stage iteratively and aggregate them for the final prediction. }
    \label{fig:main_pic}
    \vspace{15pt}
\end{center}%
}]
\vspace{10pt}



\begin{abstract}
  Video action recognition (VAR) is a primary task of video understanding, and untrimmed videos are more common in real-life scenes. Untrimmed videos have redundant and diverse clips containing contextual information, so sampling the dense clips is essential. Recently, some works attempt to train a generic model to select the $N$ most representative clips. However, it is difficult to model the complex relations from intra-class clips and inter-class videos within a single model and fixed selected number, and the entanglement of multiple relations is also hard to explain. Thus, instead of ``only look once", we argue ``divide and conquer" strategy will be more suitable in untrimmed VAR. Inspired by the speed reading mechanism, we propose a simple yet effective clip-level solution based on skim-scan techniques. Specifically, the proposed Skim-Scan framework first skims the entire video and drops those uninformative and misleading clips. For the remaining clips, it scans clips with diverse features gradually to drop redundant clips but cover essential content. The above strategies can adaptively select the necessary clips according to the difficulty of the different videos.

  To trade off the computational complexity and performance, we observe the similar statistical expression between lightweight and heavy networks. Thus, it supports us to explore the combination of them. Comprehensive experiments are performed on ActivityNet and mini-FCVID datasets, and results demonstrate that our solution surpasses the state-of-the-art performance in terms of accuracy and efficiency.
\end{abstract}

\section{Introduction}


Today, kinds of videos have occupied an increasing amount of users' time with the proliferation of video applications. It has aroused significant interest in video understanding research. As the first step in video understanding, video action recognition (VAR) is responsible for classifying each video into a pre-defined category. Due to its critical role, there is a large body of research devoted to this task (e.g.,~\cite{x3d2020,feichtenhofer2019slowfast,hara2018can, tran2015learning,qiu2017learning,tran2018closer}). Most of them employ various powerful 3D Convolution Neural Networks (CNNs)~\cite{tran2018closer, hara2018can} to extract all clip's features and aggregate them for categorization. Since the untrimmed videos often last several minutes or longer and contain much redundant and diverse information in most real-life applications (e.g., on YouTube), the computational costs to process the entire videos are extremely high. Consequently, It is essential and promising to select as few representative frames~\cite{wu2019adaframe, wu2019multi, fan2018watching, wu2019liteeval,meng2020ar} or clips~\cite{gao2020listen, korbar2019scsampler} as possible to determine the category. However, there are no labels of key frames or clips. Thus, it is quite challenging to select clips from long videos, containing lots of \emph{redundant, irrelevant, and misleading information}, in a unsupervised way. Moreover, the length and information content of different videos varies greatly.

Generally speaking, the performance of clip-level selection strategies outperforms that of frame-level selection strategies because clips consisting of several consecutive frames, preserve more temporal contexts about the activities. The existing clip-level selection strategies~\cite{korbar2019scsampler,gao2020listen} try to train a neural network to learn how to select the $N$ most salient clips from dense clips. 
Moreover, the main computational costs of the VAR framework are in the heavy 3D CNNs. Thus, they often train a lightweight model to get the rough feature of each clip first, put them into the proposed selection model to select the representative clips. Then, they input the selected clips into the heavy model to get the high-quality features before obtaining the final results. 

After making a comprehensive study on the features of untrimmed videos, we rethink the limitations in the existing methods. First, from the clip-level selection strategy, key clips selection without labels is an unsupervised process. \emph{Only a single network} is hard to overcome \emph{a wide range of challenges}, like irrelevancy, interference, and redundancy. Second, from the aspect of the overall collaborative strategy of the VAR framework, if the weaker features obtained from a lightweight model are used to select representative clips, it will further harm the ability of the selection model to select high-quality clips. Third, in terms of training strategy, some untrimmed datasets~\cite{caba2015activitynet} annotate temporal boundaries where action instances related to the action label. The original works only use the instances within boundaries to train clip-level classifiers and selection networks. However, they need to test  all clips from the whole videos, resulting in a certain shift bias between the training and testing sets, and leading to degrading results in the final test. Last, the selection network lacks interpretability in the black box, and it may suffer from data bias causing poor generalization ability in unseen activities.

To address the above issues, in this paper, we try to decouple~\cite{hsu2017unsupervised, locatello2020commentary, split2020} different aspects of challenges in a ``divide and conquer" strategy. In terms of selection strategy, we propose a simple yet effective clip-level VAR solution based on skimming and scanning techniques, which are originally proposed to facilitate people to achieve quick yet effective reading~\cite{maxwell1972skimming}. As shown in Figure~\ref{fig:main_pic}, we take an information theory-guided approach to achieve automated skimming and scanning. In specific, first, we observe that the size of the \emph{entropy is strongly correlated with the proportion of positive and negative clips}. The smaller the entropy, the larger the proportion of positive clips; therefore, we cam \textbf{skim} by a simple entropy threshold to exclude some uninformative clips. Second, we observe that untrimmed video has many different action classes in contextual scenes, and such clips are highly misleading. Meanwhile, thanks to the annotation of the temporal boundaries, we use these labels to supervise a lightweight class discriminator to distinguish the annotated and unannotated clips, i.e., to train a class-dependent binary classifier to \textbf{skim} the clips that interfere with the current class. Furthermore, some datasets~\cite{jiang2017exploiting} and real-world scenes do not have the binary annotation, which is time-consuming and labor-intensive to be labelled. It motivates us to further explore the transfer capability of the proposed class discriminator.
Third, we observe that there are still many clips with similar appearance in the left clips, and their information gain is small. Meanwhile, we observe that the existing works tend to miss some key clips. Inspired by the idea of information gain~\cite{kullback1951information}, we can \textbf{scan} the clips that do not repeat and also cover the necessary content of the video as much as possible.
In summary, the above Skim-Scan strategy show good interpretability, well generalization ability, and can adaptively select clips which is more reasonable of untrimmed videos.

Besides, the trade-off between performance and computational costs is necessary to be better considered. Unlike the motivation of previous approaches~\cite{korbar2019scsampler,gao2020listen}, we do so by the observation that the statistical expression of features between lightweight and heavy models are almost approximate based on our information theory-guided approach. It guarantees that the feature of the lightweight models can play an effective role in the selection stage. Hence, we extend our Skim-Scan Selection framework into the Slim-Scan framework.

Finally, we conduct comprehensive experiments on ActivityNet-v1.3~\cite{caba2015activitynet} and mini-FCVID~\cite{jiang2017exploiting} to validate the performance of our Skim-Scan Framework and Slim-Scan Framework. Compared to state-of-the-art techniques~\cite{gao2020listen, korbar2019scsampler}, 
our framework on average decrease 30\% clips number with the comparable accuracy in ActivityNet1.3~\cite{caba2015activitynet} and mini-FCVID~\cite{jiang2017exploiting} in both two frameworks. Furthermore, our framework gains 1.0\%-2.9\% and 0.5\% - 1.0\% improvement on accuracy in ActivityNet1.3~\cite{caba2015activitynet} and  mini-FCVID~\cite{jiang2017exploiting} respectively with the same computation costs in these two settings. 

\section{Related Work}

\begin{figure*}[!ht]
\centering 
\includegraphics[width = 15cm]{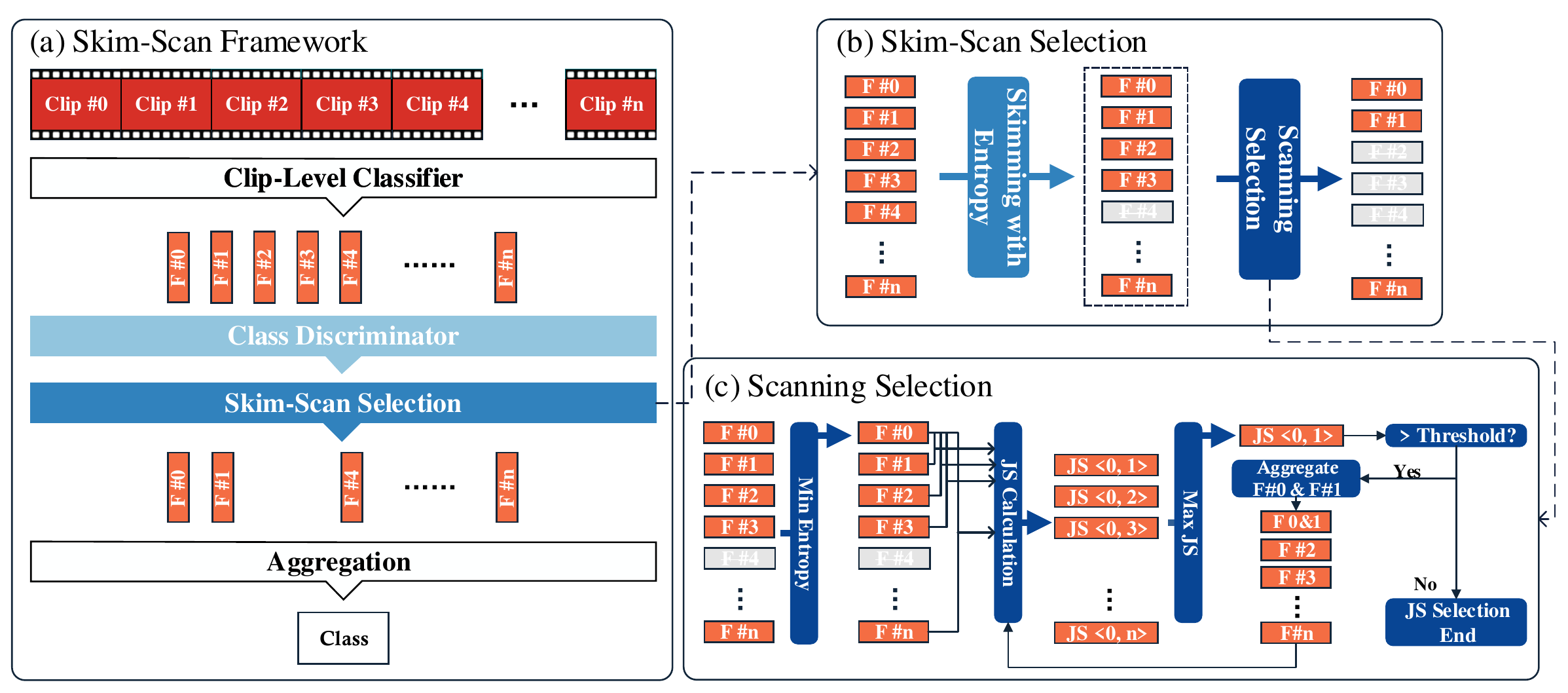}
\caption{(a) shows the workflow of the proposed Skim-Scan selection framework, which consists of a class discriminator to drop misleading clips, skimming with entropy threshold to decrease the uninformative clips, and scanning with JS divergence to drop the redundant clips and maintain the essential content. Finally, aggregation these selected clips' features ($3\%\sim5\%$) to obtain the final video-level prediction.}
\label{fig:workflow}
\end{figure*}


The objective of video action recognition task is to classify videos into a predefined set of classes. Early works employ 2D CNN models~\cite{he2015deep, tan2019efficientnet} for VAR. Later VAR frameworks utilize 3D CNNs~\cite{hara2017learning, tran2018closer,tran2015learning,lea2017temporal,hara2018can} to extract the features from the video clips because the clips convey the Spatio-temporal information, which is crucial for video understanding. 
Mainstream clip-level VAR techniques \cite{feichtenhofer2019slowfast,x3d2020,tran2015learning,hara2017learning, kataoka2020would} is consisted of three steps.
\begin{enumerate}
    \item Divide the long video into a set of short fixed-length video clips without overlapping with each other.
    \item Apply clip-level classifier to all the clips to get their classes.
    \item Aggregate all the clip-level predictions to obtain the video-level prediction. 
\end{enumerate}

\vspace{-0.1cm}

Previous works~\cite{hara2017learning, tran2015learning,lea2017temporal} incur extremely high computation costs due to the heavy 3D CNN models and the tremendous amount of input data. recent works~\cite{tran2018closer,zolfaghari2018eco,feichtenhofer2019slowfast,x3d2020, tran2019video} introduce various efficient 3D CNN models to reduce the model complexity. Meanwhile, different sampling strategies ~\cite{fan2018watching,wu2019adaframe,meng2020ar,wu2019liteeval,wu2019multi,korbar2019scsampler} propose to reduce the redundant clips to improve efficiency and performance. The sampling techniques can be divided into two categories.


\vspace{-0.2cm}
\paragraph{\textbf{Frame-level sampling}} 
These strategies normally deploy a LSTM model~\cite{fan2018watching, wu2019adaframe, wu2019liteeval, wu2019multi,meng2020ar} to look ahead the video stream and select limited number of frames~\cite{he2015deep,tan2019efficientnet,karpathy2014large}. Only the selected frames are fed into the subsequent classifier and aggregator,   which results in approximately $50\% - 60\%$ GFLOPs computation reduction~\cite{wu2019adaframe, wu2019liteeval,meng2020ar}. These strategies propose black-box algorithms, such as reinforcement learning(RL)~\cite{fan2018watching,wu2019adaframe}, multi-agents reinforcement learning~\cite{wu2019multi}, and Gumbel Softmax Sampling~\cite{meng2020ar, wu2019liteeval} to train the selector. These strategies, however, have two shortcomings. First, the \emph{Spatio-temporal information is missing} in the selected frames, which is essential for action recognition~\cite{ji20123d, hara2017learning}. Second, these \emph{black-box training algorithms have limited interpretability}~\cite{puiutta2020explainable}. 


\vspace{-0.2cm}
\paragraph{\textbf{Clip-level sampling}} 
Clip-level features obtain the Spatio-temporal information, and recent works~\cite{korbar2019scsampler, gao2020listen} reveal the superiority. \emph{Scsampler}~\cite{korbar2019scsampler} proposes to train a lightweight 2D CNN model to determine the ``saliency score'' of all clips and then select the $N$ most salient clips for subsequent classification. Consequently, \cite{gao2020listen} replaces the costly analysis of a clip with a single frame and its accompanying audio to reduce short-term temporal redundancy. It also employs an attention-based LSTM to select the most similar clips in the embedding space iteratively. For instance, they average features from the entire clips of the video to obtain the first clip, then select the second clip with maximal attention score among left clips. Unfortunately, we observe that the selected clips contain visual redundancy in both methods. As a result, \emph{valuable but less salient clips may be excluded}, which leads to a misunderstanding of the video's complete content, \emph{if only a fixed number of clips are selected.} Moreover, without externally considering the interference of \emph{redundant, irrelevant, and misleading} clips in untrimmed videos, these methods try to solve all challenges in a model without corresponding supervision, making the learning process hard.


Based on the above limitations, we propose an Skim-Scam framework to select clips step by step, which is effective, model-agnostic, and better in terms of interpretability.

\section{Approach}

Before discussing the details of the proposed approaches, we first explain our observation and motivation (Sec.\ref{sec:method1}). In Sec.\ref{sec:method2}, we design a \emph{Skim-Scan} framework to pick a few most representative clips after extracting all clips' features with the help of the clip-level classifier. In Sec.\ref{sec:method3}, we further extend our framework by substituting the costly clip-level classifier with two collaborative classifiers. We denote this framework as \emph{Slim-Scan}.

\subsection{Observation and Motivation}
\label{sec:method1}


\begin{figure}[!h]
\centering 
\includegraphics[width = 8.6cm]{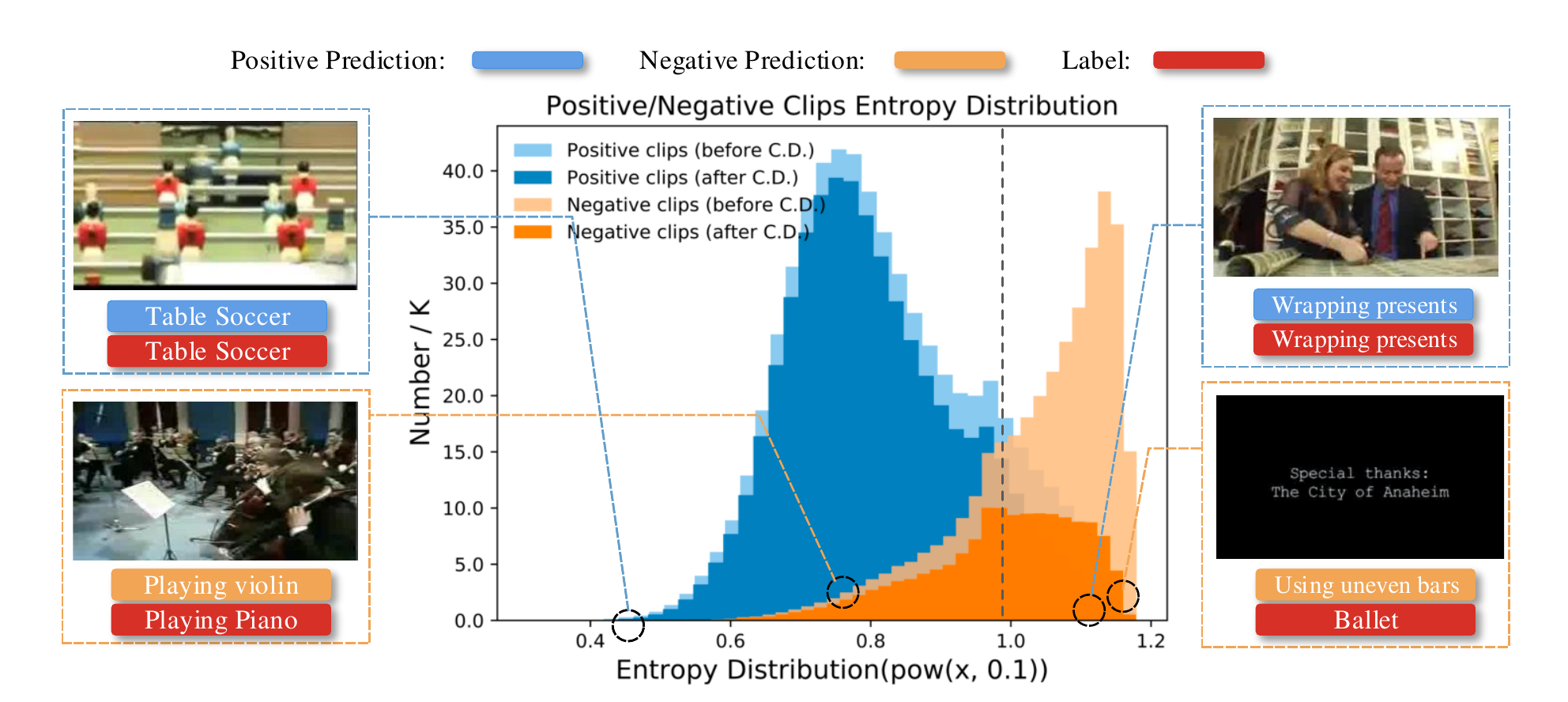}
\caption{Entropy distribution of the feature of clips and four visualization clips with labels, indicating four cases of positive/negative clips in low/high entropy. Positive means the clip's label is the same as the video's label, and vice versa. C.D. is an abbreviation for Class Discriminator.}
\label{fig:entropy_sampler}
\end{figure}




\paragraph{\textbf{Motivation of skimming with entropy to drop uninformative clips}} 
Entropy is always used to measure the uncertainty of information in Information Theory~\cite{robinson2008entropy}. Thus, we propose to use the entropy of the clip's classification probability to measure the classification uncertainty of each clip. The larger the entropy, the higher the uncertainty of the classification, which means that the clip is more uninformative. Further, we analyze the statistic of the classification entropy~(defined in Sec.~\ref{sec:method2}) distribution of positive clips and negative clips in Figure~\ref{fig:entropy_sampler}. We find that when entropy is small, the number of positive clips far exceeds negative clips. As entropy increases, the proportion of negative clips increases and exceeds positive clips around a certain value~(dashed line). Therefore, it motivates us to skim uninformative clips based on a proper threshold of entropy.

\paragraph{\textbf{Motivation of the transferable Class Discriminator to drop misleading clips in the untrimmed videos}}
Within one untrimmed video, there may be more than one activity instance from more than one activity class~\cite{caba2015activitynet}. These clips may be similar to other inter-class videos, so they have a sufficient amount of information. However, they do not reflect the content of the ground-truth label of this video, such as the lower-left corner of the picture in Figure~\ref{fig:entropy_sampler}, which brings strong interference and misleading. Besides, some datasets of untrimmed videos are labeled as \emph{annotated clips} and \emph{unannotated clips}, which can represent whether the corresponding clip is consistent with the current video class. This information motivates us to design a class discriminator to train a class-dependent binary classifier to remove clips that interfere with the current class. Furthermore, considering that some datasets do not have this annotation information and manual annotation is time-consuming and labor-intensive, we further explored the transferability of the proposed class discriminator.


\begin{figure}[!h]
\centering 
\includegraphics[width = 8cm]{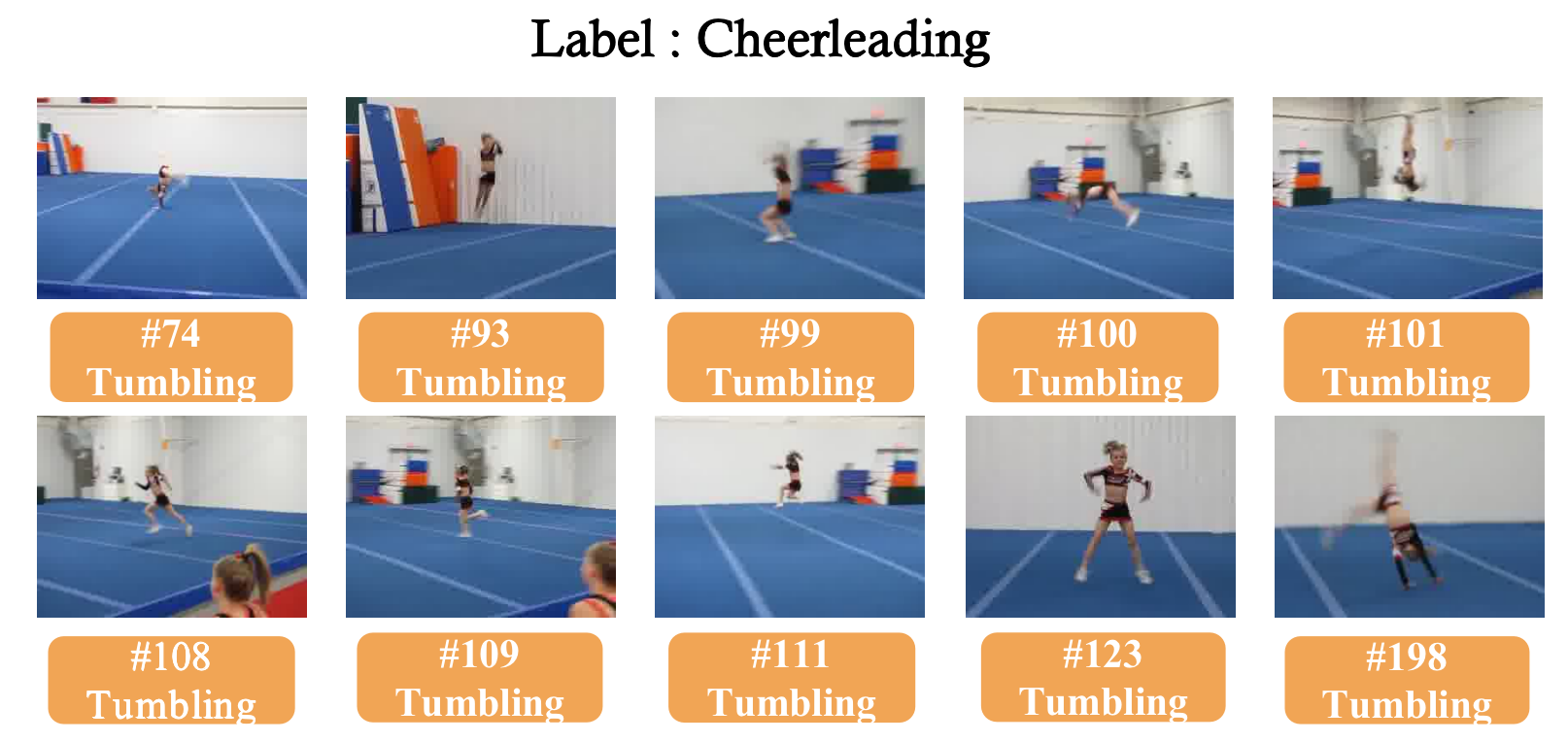}
\caption{The negative selected clips from \emph{Scsampler}.}
\label{fig:entropy_sampler_case}
\end{figure}

\paragraph{\textbf{Motivation of scanning with Information Gain to remove redundancy}}
Accordingly, there are still a large number of clips after the above skimming with redundant information. Meanwhile, we find that there are multiple diverse clips in untrimmed videos, which should not be ignored to consider the clips' relations and cover essential content in the video. For instance, in Figure~\ref{fig:entropy_sampler_case}, we find that Scsampler~\cite{korbar2019scsampler} misjudges the prediction because it only catches the attention on clips with the same "Tumbling" action (sorting top-10 salient clips) but ignores the "Cheerleading" clips. Thus,  to further reduce the number of clips needed while including as much diverse content as possible from the video, we propose to scan with information gain strategy.

\subsection{Skim-Scan Framework}
\label{sec:method2}
According to the above motivation, we suppose that the divide-and-conquer strategy is more suitable for the untrimmed videos. Here, we introduce the three stages of our Skim-Scam framework as follows.


\paragraph{\textbf{Skimming clips and drop uninformative clips based on the entropy threshold:}}
Each clip feature $V_i$ obtains the classification probability distribution $P_i$ through a clip-level classifier $f$ as the Eq.~\ref{eq:prob} shows. 

\begin{equation}
     P_i = Softmax\big(f(V_i)\big) 
\label{eq:prob}
\end{equation}

Entropy~\cite{shannon2001mathematical} measures the uncertainty of the information, which can be an indicator of the classification confidence of each clip according to $P_i$. The entropy of the classification probability distribution $H(P_i)$ is shown in Eq.\ref{eq:entropy}. When $H(P_i)$ is high, $P_i$ tends to be uniform, and thus the clip $V_i$ is usually uninformative. In contrast, when $H(P_i)$ is low, the distribution $P_i$ has a clear tendency. It means the clip has strong confidence to be in a specific class.

\begin{equation}
\centering
\begin{gathered}
    H(P_i) = -\sum_{k \in \mathcal{C}}P_{i,k}log(P_{i,k}),
\end{gathered}
\label{eq:entropy}
\end{equation}

where $\mathcal{C}$ is the set of predefined classes, and $P_{i,k}$ means the probability that the clip $i$ belongs to the class $k$.
Therefore, to eliminate useless or noise information, we drop these clips $i$ with a proper entropy threshold $\Theta_{Entropy}$ preliminarily in Eq.\ref{eq:threshold}.

\begin{equation}
H(P_i) \geq \Theta_{Entropy}
\label{eq:threshold}
\end{equation}

\paragraph{\textbf{Skimming clips and drop misleading clips with Class Discriminator}}

\begin{figure}[!h]
\centering 
\includegraphics[width = 8cm]{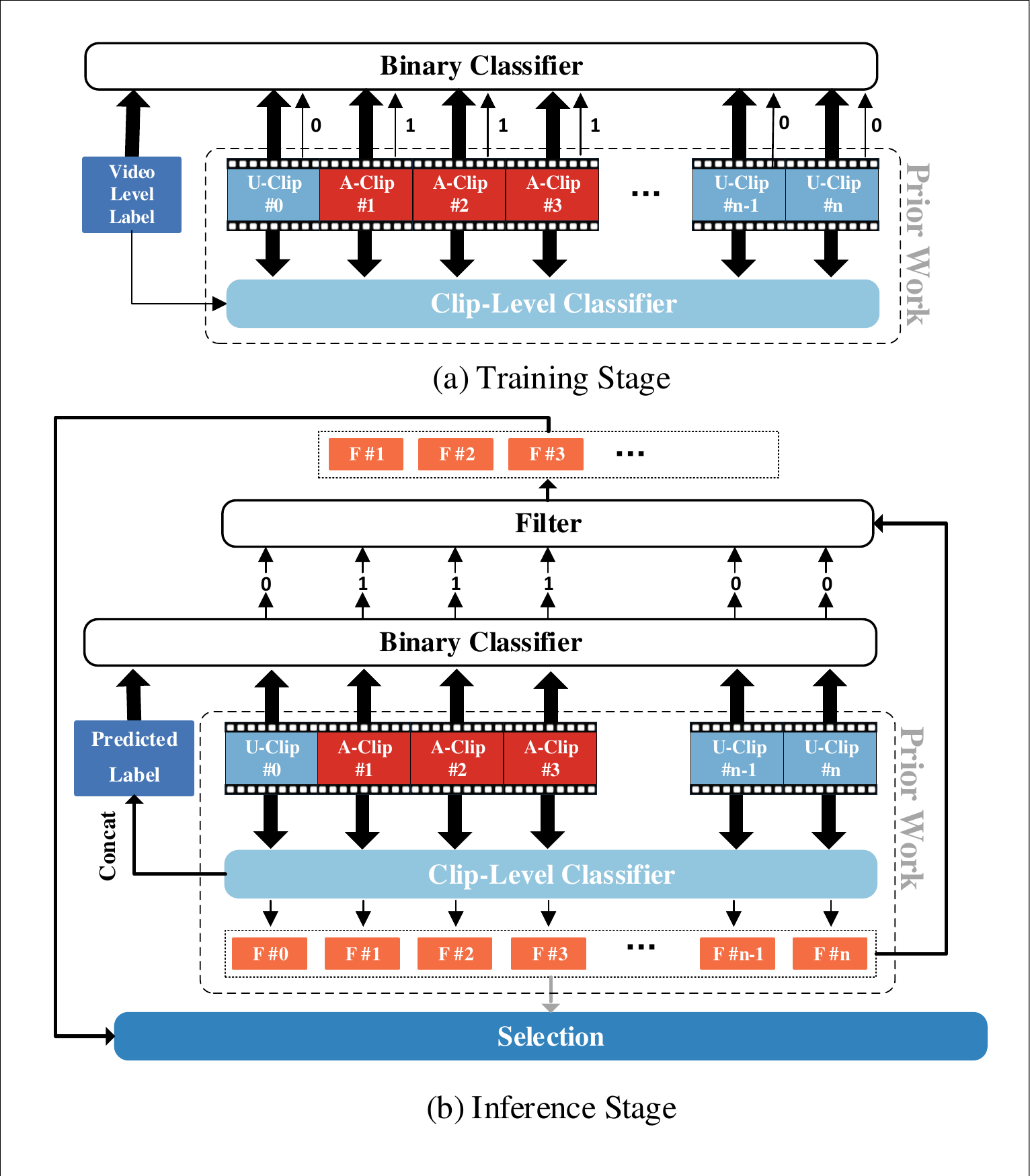}
\caption{The workflow of training and inference stages for both the clip-level classifier and the Class Discriminator. The training and inference stage of the clip-level classifier is the same as the prior work. The binary classifier distinguishes unannotated and annotated clips by inputting the concatenated the clip's features and the video-level label. In the training stage, the \emph{thick arrow} represents the inputs, and the \emph{thin arrow} represents the supervision information.}
\label{fig:binary}
\end{figure}

We propose a \emph{Class Discriminator}, shown in Figure~\ref{fig:binary}, to distinguish the unannotated clips and annotated ones, which can be marked as the class-related clips and the irrelevant clips in untrimmed videos. Moreover, the video-level label can be critical guidance to properly distinguish whether the clips are annotated. Thus, during training, we utilize a one-hot vector $V_c$ to represent the class label of the video $c$ and concatenate them with the clip's features $F_i$ to train the class discriminator. The target is to predict the type of each clip (i.e., unannotated or annotated ones) under the supervision, where those within the temporal boundaries are of class $1$ and those outside the boundaries are of class 0. In the inference stage, we first aggregate all clips' features to obtain the predicted class. Accordingly, the proposed class discriminator drops the ``unannotated'' clips to avoid these misleading clips being selected in the following procedure.



In addition, since most videos are untrimmed without any annotations, the transferability of the proposed \emph{class discriminator} will be beneficial. Specifically, on the unannotated dataset, we stitch together the class label and the clip features into the pretrained class discriminator and throw away the unwanted clips, where the prediction class is $0$. Then, we average the features from retained clips to obtain the final predicted class. Finally, we use the class label of the video as supervision to finetune the intermediate binary discriminator.
\paragraph{\textbf{Scanning to select the clips with the largest information gain iteratively:}} To cover the essential information of videos with as few clips as possible, we propose to select the clip that brings the maximum Information Gain upon the whole information of currently selected clips iteratively. Information Gain\cite{kullback1951information} measures the complementarity between information and the decrease of the uncertainty of information. That is, the greater gains from new information, the more entropy can be reduced. Due to the precise estimation of the distribution of Information Gain is quite complex, we propose to utilize the distance between clips' distribution $P_{ij}$ to approximate the Information Gain. In other words, if clip $j$ is farthest away from clip $i$, it means that clip $j$ can provide the most information to clip $i$ compared to other clips. Here we average all selected clips' features to represent currently selected information. Considering the symmetry property of selected clips and to-be-selected clip, we utilize JS Divergence~(Eq.~\ref{eq:js}) to substitute KL Divergence~(Eq.~\ref{eq:kl}) to approximate the Information Gain. We further discuss other metrics to approximate the information gain in the  \emph{Supp}.
\begin{equation}
\centering
D_{KL}(P_i||P_j) = \sum_{k \in \mathcal{C}} P_{i,k} log( \frac{P_{i,k}}{P_{j,k}})
\label{eq:kl}
\end{equation}
\begin{equation}
\centering
D_{JS}(P_i||P_j) = \frac{1}{2}D_{KL}(P_i||\frac{P_i+P_j}{2}) + \frac{1}{2}D_{KL}(P_j||\frac{P_i+P_j}{2})
\label{eq:js}
\end{equation}

The Scanning stage consists of three steps as Figure~\ref{fig:workflow}(c). In step $1$, we select the clip with the minimum entropy as the first selected clip (e.g., $F \#0$). In step $2$ (Eq.~\ref{con:6}), we calculate the JS Divergence of prediction distributions between each unselected clip $P_j$ and the aggregation of the all selected clips $I_\mathcal{S}$~(e.g., $F \#0$). We select the clip (e.g., $F \#1$) maximizes the JS divergence. 


\begin{equation}
j = \mathop{\arg\max}_j D_{JS}(I_\mathcal{S}, P_j)
\label{con:6}
\end{equation}

In step 3 (Eq.~\ref{con:5}), we update $I_\mathcal{S}$ by taking the new selected clip into consideration~(e.g., $F \#0 ~\& ~F\#1$). 

\begin{equation}
\centering
\begin{aligned}{
I_\mathcal{S} = Softmax(\frac{1}{|\mathcal{S}|} \sum_{i \in \mathcal{S}} f(V_i)),
   }
\end{aligned}
\label{con:5}
\end{equation}
where $\mathcal{S}$ is the set of selected clips.
We repeat step $2$ and $3$ to iteratively select more clips until the largest JS divergence is less than a threshold ($0.4$ by default) in step $2$.

\begin{figure}[!h]
\centering 
\includegraphics[width = 7cm]{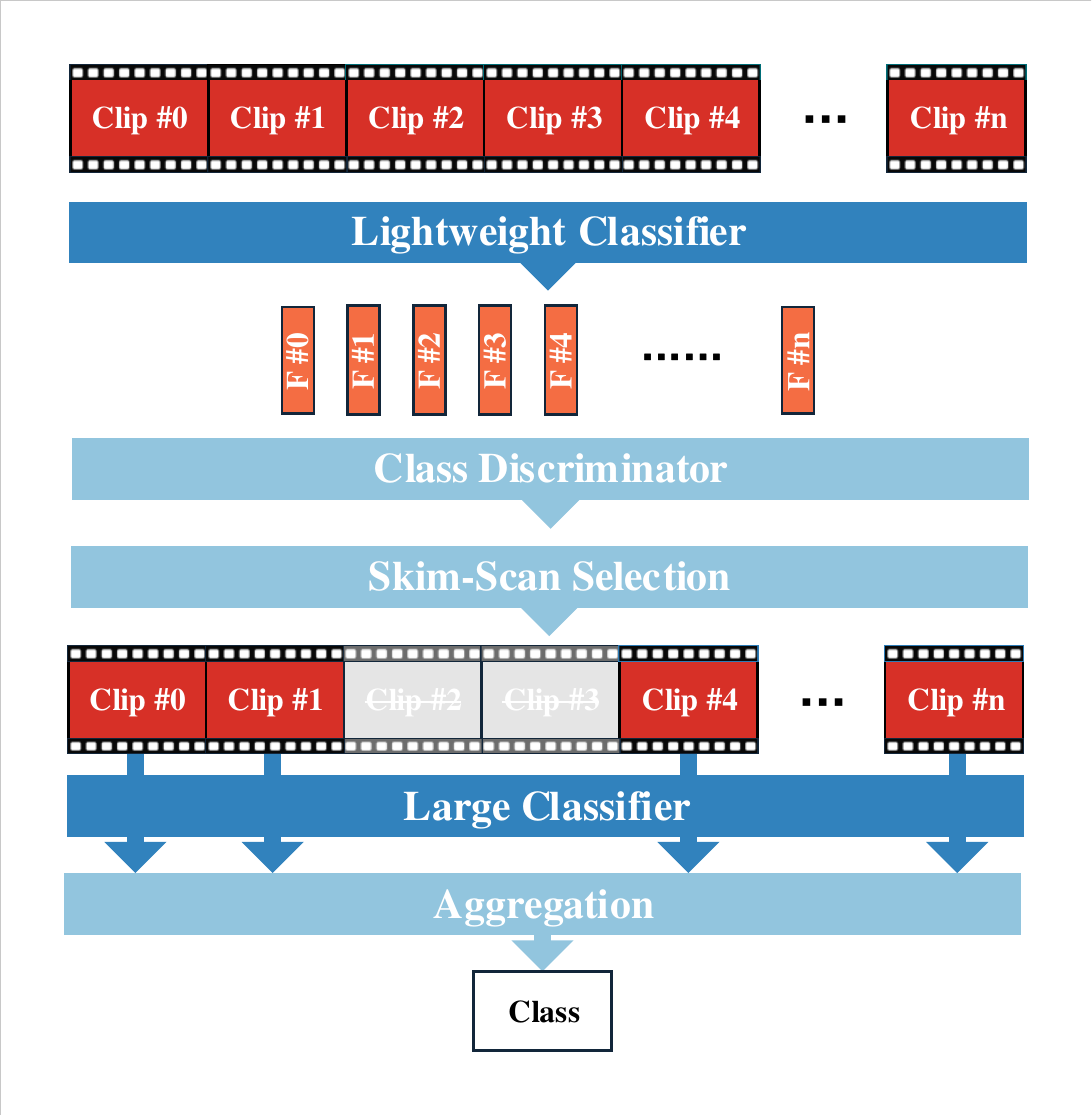}
\caption{The framework of the Slim-Scan Framework. A lightweight classifier first extract the primary features of all clips, putting them into the proposed Skim-Scan Selection framework to select the representative clips. Last, a large classifier recalculate better features for the selected clips to get the final video-level prediction.}
\label{fig:Slim-Scan}
\end{figure}

\subsection{Slim-Scan Framework}
\label{sec:method3}
Because the mainly computational costs are in the clip-level classifier, and considering we utilize the statistic information (i.e., Entropy and JS Divergence) of clips' features in the selection stage, rather than the clips' features itself~\cite{korbar2019scsampler,gao2020listen}, we propose that a lightweight network has the capability to approximate the heavy network in a similar statistic expression.

Consequently, we extend our Skim-Scan Selection Strategy to a Slim-Scan Framework. As Figure~\ref{fig:Slim-Scan} shows, we first utilize a lightweight classifier, such as ShuffleNet3D\cite{kopuklu2019resource}, to preliminary extract the clip-level features of all clips, and we select the clips based on the Skim-Scan selection strategy. Only the selected clips (round $3\% \sim 5\%)$ will be recalculated by a heavy classifier~\cite{tran2018closer,hara2018can}. We then aggregate the more accurate features to obtain the final video-level predictions.

Due to the limited performance of a lightweight model, we resort to the standard knowledge distillation technique~\cite{hinton2015distilling} to boost its performance, the whole loss function $L_{dis}$ is shown in Eq.~\ref{eq:dis}. 

\begin{equation}
    L_{dis} = L_{CE} + \alpha T^2 L_{KL}(\frac{exp(\frac{t_i}{T})}{\sum_i exp(\frac{t_i}{T})}, s_i),
    \label{eq:dis}
\end{equation}
wherein $L_{CE}$ is the loss of Cross Entropy, $L_{KL}$ is the loss of KL Divergence. $T$ is the temperature, $\alpha$ is the scale of KL loss, and $t_i$ and $s_i$ are the output features of the teacher network and student network, respectively. 

\section{Experiments}

\begin{table*}[t]
\centering
\renewcommand{\multirowsetup}{\centering}
\resizebox{1.02\textwidth}{19.6mm}{
\begin{tabular}{c|ccccc|ccccc|ccccc}
\hline
\multirow{3}{*}[-1.2ex]{Approach} & \multicolumn{10}{c|}{ActivityNet1.3}   & \multicolumn{5}{c}{mini-FCVID}      \\
\cline{2-16} & \multicolumn{5}{c|}{\textbf{R(2+1)D-152}}   & \multicolumn{5}{c|}{\textbf{R3D-18}} & \multicolumn{5}{c}{\textbf{R3D-18}}\\ 
\cline{2-16} & Acc & mAP & \begin{tabular}[c]{@{}c@{}}Clip \\ Num\end{tabular} & B\_Gflops & S\_Gflops & Acc  & mAP  & \begin{tabular}[c]{@{}c@{}}Clip \\ Num\end{tabular} & B\_Gflops & S\_Gflops &  Acc & mAP & \begin{tabular}[c]{@{}c@{}}Clip \\ Num\end{tabular} & B\_Gflops & S\_Gflops  \\ \hline
Dense~& 83.5 & 88.9 & 205.6 & 19.1 * 205.6 & 0.0 & 72.7 & 81.5 & 205.6 & 8.3 * 205.6 & 0.0 & 72.3 & 80.55 & 292 & 8.3 * 292.0 & 0.0     \\ 
Uniform~10 & 82.5 & 88.0 & 10.0 & 19.1 * 10.0 & 0.0 & 70.9 & 80.2 & 10.0 & 8.3 * 10.0 & 0.0 & 71.4 & 79.8 & 10.0 & 8.3 * 10.0 & 0.0 \\ 
Random~10  &  81.2 & 86.9 & 10.0 & 19.1 * 10.0 & 0.0 & 69.0 & 78.6 & 10.0 &  8.3 * 10.0 & 0.0 & 70.4 & 79.1& 10.0 & 8.3 * 10.0 & 0.0 \\ 
Scsampler~10 & 84.0 & 88.6 & 10.0 & 19.1 * 205.6 & 0.0 & 74.5 & 82.7 & 10.0 & 8.3 * 205.6 & 0.0  & 72.0 & 80.1& 10.0 &  8.3 * 292.0 & 0.0   \\
Listen to Look~10  &  79.6 & 85.6 & 10.0 & 19.1 * 205.6 & 0.37 & 74.7  & 82.9 & 10.0 & 8.3 * 205.6 & 0.14 & 72.6 & 80.7 & 10.0 & 8.3 * 292.0 & 0.16  \\ \hline
Skim-Scan~(+) &  84.2 & 89.5 & \textbf{2.9} & 19.1 * 205.6 &  0.012 & 74.8 & 83.0 & \textbf{6.5} & 8.3 * 292.0 & 0.003 & 72.6 & 80.8 & \textbf{7.2} & 8.3 * 292.9 & 0.004  \\ 
Skim-Scan~(10)  &  \textbf{86.4} & \textbf{90.7} & 9.2 & 19.1 * 205.6 & 0.012 & \textbf{75.8} & \textbf{83.7} & 10.6 & 8.3 * 292.0 & 0.003 & \textbf{73.3} & \textbf{81.3} & 10.1 & 8.3 * 292.0 & 0.004 \\
\hline
\end{tabular}}

\vspace{0.1cm}
\caption{Comparison different sampling strategies of video-Level prediction on ActivityNet-v1.3~(\#classes:200) and mini-FCVID~(\#classes:239) datasets, based on the same clip-level classifiers. Higher values in Acc (Accuracy \%) and mAP are better. Lower values in Clip Num, B\_Gflops (the Gflops per video of the Backbone), and S\_Gflops (the Gflops per video of the Selection strategies) are better. Best results in bold. Skim-Scan~(10) means to select nearly 10 clips. Skim-Scan~(+) means to select proper clips to reach the sota.}
\label{table:Main Table1}
\end{table*}

\begin{table*}[t]
\centering
\renewcommand{\multirowsetup}{\centering}
\resizebox{1.02\textwidth}{12mm}{
\begin{tabular}{c|ccccc|ccccc|ccccc}
\hline
\multirow{3}{*}[-1.2ex]{Approach} & \multicolumn{10}{c|}{ActivityNet1.3}   & \multicolumn{5}{c}{mini-FCVID}      \\
\cline{2-16} & \multicolumn{5}{c|}{\textbf{Shufflenet-V1/R(2+1)D-50}}   & \multicolumn{5}{c|}{\textbf{Shufflenet-V2/R(2+1)D-50}} & \multicolumn{5}{c}{\textbf{Shufflenet-V2/R(2+1)D-50}}\\ 
\cline{2-16} & Acc & mAP & \begin{tabular}[c]{@{}c@{}}Clip \\ Num\end{tabular} & B\_Gflops & S\_Gflops & Acc  & mAP  & \begin{tabular}[c]{@{}c@{}}Clip \\ Num\end{tabular} & B\_Gflops & S\_Gflops &  Acc & mAP & \begin{tabular}[c]{@{}c@{}}Clip \\ Num\end{tabular} & B\_Gflops & S\_Gflops  \\ \hline
Scsampler 10 &  78.5 & 85.8 & 10 & 0.39 * 205.6 + 10.6 * 10.0 & 0.0 &  78.3 & 85.5 & 10 & 0.36 * 205.6 + 10.6 * 10 & 0.0 & 74.4 & 82.4 & 10.0 &  10.6 * 10.0 & 0.0 \\
Listen to Look 10  & 80.3 & 87.0 & 10 & 0.39 * 205.6 + 10.6 * 10.0 & 0.35 &  80.2 & 87.1 & 10 & 0.36 * 205.6 + 10.6 * 10 & 0.37 &  77.9 & 84.8 & 10.0 & 0.36 * 292 + 10.6 * 10.0&  0.46    \\ \hline
Slim-Scan~(+) & 80.3 & 87.0 & \textbf{7.0} & 0.39 * 205.6 + 10.6 * 7.0 & 0.012 & 80.2 & 87.1 & \textbf{6.9} & 0.36 * 205.6 + 10.6 * 6.9 & 0.012 & 77.9 & 84.9 & \textbf{6.8} & 0.36 * 205.6 + 10.6 * 6.8 & 0.004   \\ 
Slim-Scan~(10)  & \textbf{81.0} & \textbf{87.6} & 9.8 & 0.39 * 205.6 + 10.6 * 9.8 & 0.012  & \textbf{81.0} & \textbf{87.6} & 9.7 & 0.36 * 292.0 + 10.6 * 9.7 & 0.012 & \textbf{78.4} & \textbf{85.2} & 9.4 & 0.36 * 292.0 + 10.6 * 9.4 & 0.004  \\ \hline
\end{tabular}
}
\vspace{0.1cm}
\caption{Comparison different sampling strategies of Video-Level prediction on ActivityNet-v1.3 and mini-FCVID datasets, based on the same clip-level feature from \emph{combination of lightweight backbones and  heavy backbones}.}
\label{table:Main Table2}
\end{table*}

\paragraph{\textbf{Dataset}}
\textbf{ActivityNet-v1.3}~\cite{caba2015activitynet} is an untrimmed video dataset labelled with 200 action classes and contains 10024 videos for training and 4926 videos for validation with an average duration of 117 seconds. They manually annotate the temporal boundaries where activity instances are associated with exactly one ground truth activity label. 
\textbf{Mini-FCVID}~\cite{jiang2017exploiting} is another untrimmed video dataset containing randomly selected 22045 videos (11468 videos for training and 11577 videos for testing) with 239 categories, and the average length is 167 seconds. There are no labels for the temporal boundaries of each activity instance. 

\paragraph{\textbf{Implementation Details}}

We use R(2+1)D-152~\cite{tran2018closer}, R(2+1)D-50~\cite{tran2018closer}, R3D-18~\cite{hara2018can}, Shufflenet-V1~\cite{kataoka2020would}, Shufflenet-V2~\cite{kataoka2020would} as the different clip-level classifiers. We can use the annotations in ActivityNet-v1.3~\cite{caba2015activitynet} to train the proposed class discriminator in a supervised way. Meanwhile, the structure of the class discriminator is: Conv1d (1, 1, 1) $\rightarrow$ Global Average Pooling (channel size) $\rightarrow$ Fully Connection Layer(channel size, 2), which brings tiny computer overhead compared with the clip-level classifiers. The channel size depends on the output size of the clip-level classifier. We set the channel size as $2000$ for $2048$ output size and $500$ for $512$ output size. The supervised training learning rate is $0.005$, with SGD optimizer. The epoch is 4. And the transfer training learning rate is 0.001, with SGD optimizer. The epoch is $80$. For the threshold of entropy and JS Divergence, we set the entropy threshold to retain roughly $60\%$ of the clips and set JS Threshold=0.5 by default.
For the loss of slim-scan framework,  we set $\alpha$=0.8,  $T$=1 in Eq.~\ref{eq:dis}.


\subsection{Comparison of Clip-Level Sampling Frameworks}
\label{sec:experiment1}


We first compare the Skim-Scan Selection framework to the following baselines and several existing methods\cite{korbar2019scsampler,gao2020listen}.

\begin{itemize}

    \item DENSE: It averages the clip-level prediction scores from all clips as the video-level prediction.
    \item RANDOM: It randomly samples 10 clips out of all clips
    \item UNIFORM: The same as the ``RANDOM'' except that we perform uniform sampling.
    \item SCSAMPLER~\cite{korbar2019scsampler}: It selects the top 10 clips with the highest \emph{confidence scores}.
    \item LISTEN-TO-LOOK~\cite{gao2020listen}: The work proposes two approaches, one sampling approach and the other unsampling approach. We first compare our sampling frameworks with the sampling approach here. It iteratively selects the clip with the largest \emph{attention score} by 10 times. For the unsampling approach, we will compare it in ~\ref{sec:experiment3} individually.

\end{itemize}
At last, all methods average the clip-level predictions of these 10 clips as the video-level prediction.

In Table~\ref{table:Main Table1}, we compare the various approaches based on the same clip-level features from widely used backbones~\cite{tran2018closer,kataoka2020would}, to avoid the effect from the different feature extractors. Since previous works~\cite{korbar2019scsampler, gao2020listen} always select $10$ clips for each video, we also select 10 clips in \emph{Uniform} and \emph{Random} methods. While our strategy can select clips \textbf{adaptively}, to be fair, we adjust our thresholds to keep nearly 10 clips on average. We observe that our Skim-Scan Framework gains $2.9\%$~(accuracy from $84.0\%$ to $86.4\%$) and $1.5\%$~(accuracy from $74.7\%$ to $75.8\%$) improvement on ActivityNet-v1.3 and $1.5\%$ on mini-FCVID~(accuracy from $72.6\%$ to $73.7\%$). Furthermore, we adjust the thresholds, marked with Skim-Scan~(+), to achieve comparable accuracy with existing works. Under this setting, our method can reduce the amount of calculation by $30\%\sim70\%$.


Considering that \emph{Scsampler} and \emph{Listen to look} also supports the combination of a heavy and the lightweight networks. Therefore, we compared the selection strategies under the Slim-Scan Framework with them. 
We utilize the same combinations of the heavy and the lightweight backbones among these three approaches in Table~\ref{table:Main Table2}, which ensures only the selection strategies influence their performance.
Similar to the Table~\ref{table:Main Table1}, we first adjust our thresholds to keep nearly 10 clips on average for a fair comparison to \emph{Scsampler} and \emph{Listen to look}. In Table~\ref{table:Main Table2},  We observe that our Slim-Scan framework gains $0.9\%$~(accuracy from $80.3\%$ to $81.0\%$) and $1.0\%$~(accuracy from $80.2\%$ to $81.0\%$) improvement on ActivityNet-v1.3 and $0.6\%$ on mini-FCVID~(accuracy from $77.9\%$ to $78.4\%$). Furthermore, it also shows that our framework can reduce around $20\%$ computation costs by the comparable performance with others. The improvement of our sampling strategy indicates effectiveness and superiority.

\subsection{Ablation Study}
\label{sec:experiment2}
\paragraph{\textbf{Impact of each stage within Skim-Scan Selection framework}}
 We define seven different combinations of these stages in Table~\ref{table:strategy}. For a fair comparison, we use the same setting and backbone (R(2+1)D-152) of all strategies. 

Compared with the baseline ``Dense" strategy, our Skim-Scan Framework~(Entropy + C.D. + JS) achieves either the highest accuracy (from $83.5\%$ to $86.4\%$ in accuracy) and the least clip numbers (from 205.6 to 9.2 clips needed). The ``Entropy Only'', ``C.D. Only'' and ``Entropy + C.D.'' can gain a certain higher accuracy (by $2.0\%\sim2.9\%$), due to drop the uninformative and misleading clips. But these two stages still require quite many clips, indicating the importance of the Scanning stage with JS Divergence. Although ``JS Only'' can reduce more clips, it leads to relatively poor accuracy, because many misleading and uninformative clips cause negative effects, proving the necessity of entropy-based skimming stage and the Class Discriminator. Besides, ``Entropy + JS'' and ``C.D. + JS'' can achieve a good trade-off of accuracy and clip number, but they still have a certain gap from our full framework. It shows that the entropy skimming stage and the class discriminator have partial crossover ability and usage, and all these three parts of our framework are necessary and complementary.

\begin{table}[!h]
\centering
\resizebox{0.40\textwidth}{17.5mm}{
\begin{tabular}{c|ccc}
\hline
Strategy    & Acc. & mAP & Clip Num. \\ \hline
Dense & 83.5 & 88.9 & 205.6 \\ \hline
Entropy Only     &   85.7  &   90.2    & 117.9    \\ 
C.D. Only     &   85.2  & 90.1    & 143.4      \\ 
JS Only     &   84.8  & 89.7    & 31.2      \\ \hline
Entropy + JS &  85.8   & 90.5 &  13.0   \\ 
Entropy + C.D.     & 85.9 &  90.5  & 105.1  \\ 
C.D. +JS &   86.0 & 90.6    &  12.5   \\ \hline
Ours (Entropy + C.D. + JS) &  \textbf{86.4}   & \textbf{90.7} &  \textbf{9.2}   \\ \hline
\end{tabular}
}
\vspace{0.1cm}
\caption{Results from different stages of Skim-Scan Framework. JS and C.D. are abbreviation for JS Divergence and Class Discriminator, respectively.}
\label{table:strategy}

\end{table}

\paragraph{\textbf{Effect of the Class Discriminator}}
Our experiments show that the proposed class discriminator facilitates dropping $15\%$ positive clips and \textbf{$55\%$} negative clips in ActivityNet1.3 and $8\%$ positive clips and \textbf{$30\%$} negative clips in mini\_FCVID, significantly decreasing the proportion of negative clips. 
Additionally, the pretrained class discriminator is robust to different clip-level classifiers, which can always reach around $73.8\%$ of the binary accuracy under supervision. Meanwhile, it can further boost other selection strategies, improving the accuracy of $0.6\%$ in our strategies and $2.0\%$ in \emph{Dense} in ActivityNet1.3. Furthermore, to verify the transfer capability of the class discriminator, we compared the performance of the class discriminator with/without the pretrain model. Under the same model structure and training procedure, the class discriminator without the pretrain model shows no improvement. In contrast, with the pretrain model, we gain $0.5\%$ improvement in our framework and $0.7\%$ in \emph{Dense} in mini\_FCVID dataset.


\paragraph{\textbf{Impact of the threshold of Entropy and JS Divergence}}

Our strategy is dependent on the entropy and JS threshold, thus we explore the relations between the \emph{Entropy Threshold}, the \emph{JS Threshold} and the \emph{corresponding accuracy} as well as the \emph{clip numbers}.

\begin{figure}[htbp]\centering                                      
\subfigure[Heatmap of Accuracy]{                    
\begin{minipage}{3.5cm}\centering \includegraphics[width=3.4cm]{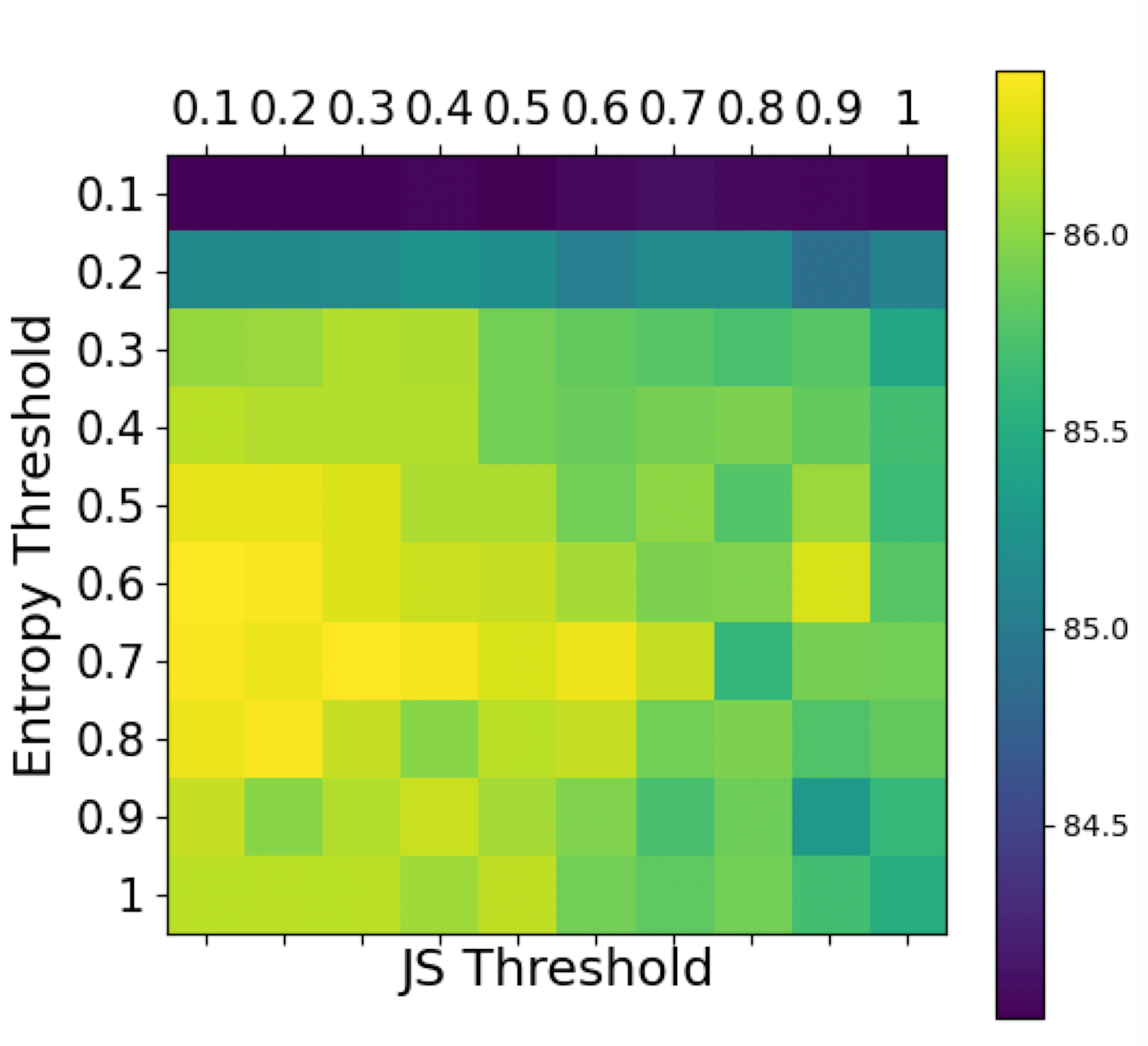}     
\end{minipage}}
\subfigure[Heatmap of Clip Numbers]{
\begin{minipage}{3.5cm}\centering \includegraphics[width=3.4cm]{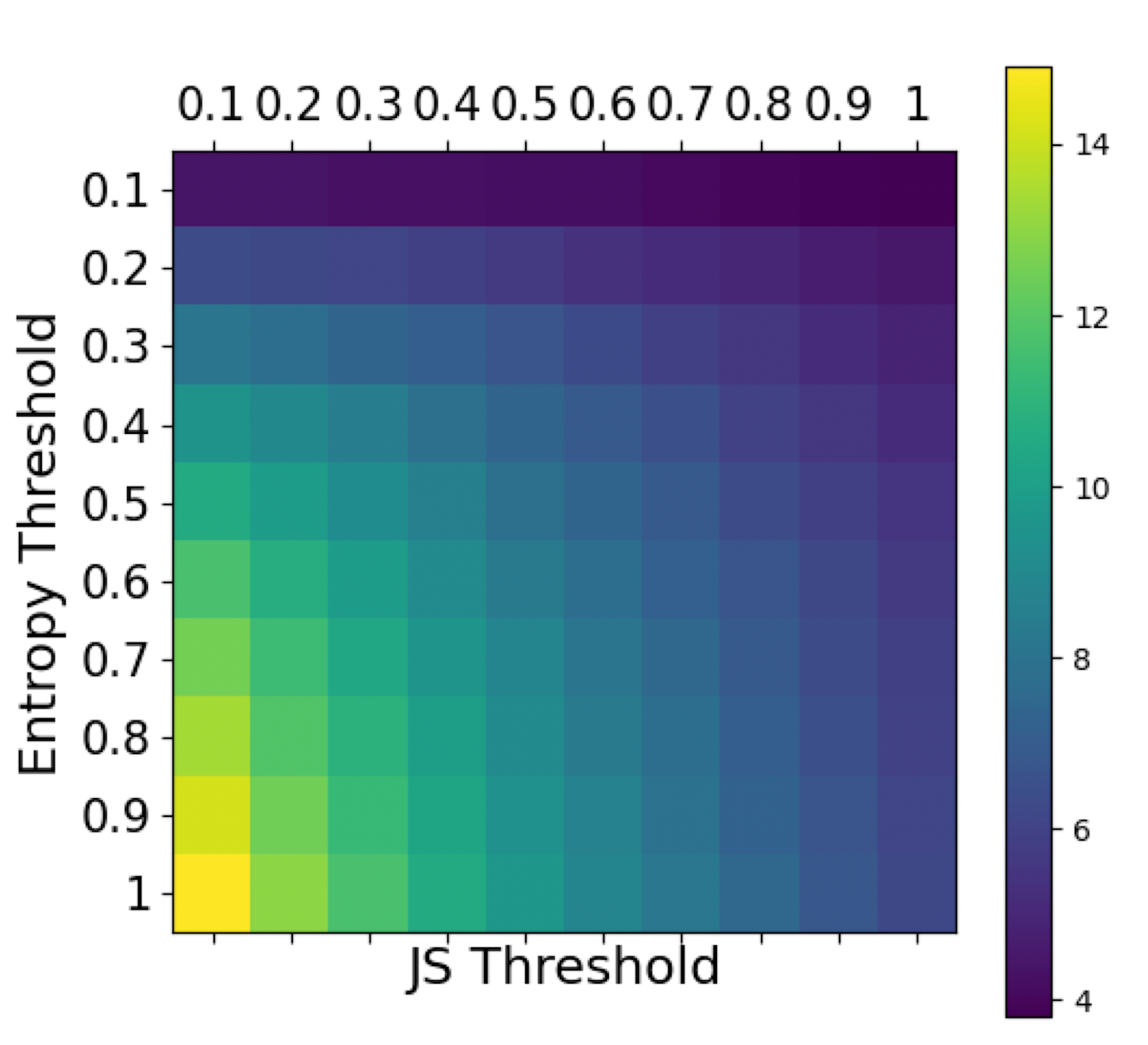}
\end{minipage}}

\caption{Heatmap of accuracy and clip number changing with Entropy threshold and JS threshold.}                       
\label{fig:threshold}                                              

\end{figure}

In Figure~\ref{fig:threshold}(a), \emph{entropy threshold} measures the proportion of informative clips of the videos. As the entropy threshold increases under the same JS threshold, accuracy first increases and then decreases slightly, reaching the best for most cases when the threshold is around $0.7$ (also shown in the vertical line in Figure~\ref{fig:entropy_sampler}), where it retains roughly $60\%$ of the clips. Such a setting achieves good performances for all the datasets used in our experiments. This stage can drop many uninformative and irrelevant clips, verifying the advantages of our skimming with entropy threshold. Meanwhile, as the JS threshold increases under the same entropy threshold, the accuracy is relatively stable. We can observe a tiny increase at the beginning and then decrease slightly. Hence, we set $\emph{JS Threshold}$=0.5 by default. It proves that we do not need all the informative clips but scan properly representative clips. 

Figure~\ref{fig:threshold}(b) indicates that the selected clip numbers and entropy thresholds are positively correlated, while the JS threshold is negatively correlated. To conclude, the entropy threshold determines the accuracy range in a coarse-grained manner, while the JS threshold is a more fine-grained tuning knob to obtain higher performance.

\paragraph{\textbf{Impact of different combination of Slim-Scan Framework}}
We further discuss the impact of the various lightweight classifiers in our Slim-Scan framework on ActivityNet-v1.3. As the Table~\ref{fig:slim-scan} demonstrates, we take five lightweight classifiers Only and R(2+1)D-50 heavy classifier Only as baseline settings. The proposed Slim-Scan framework of different combinations can reach the competitive results ($88.5$ mAP v.s. $87.7$ mAP, and $82.2\%$ v.s. $81.2\%$ in accuracy) with $10\times$ fewer flops ($1435.2$ Gflops v.s. $141.6$ Gflops). In our experiments, such statistical information calculated based on the output features of the lightweight models and the heavy models is similar. Hence, the Slim-Scan framework can make a good trade-off between accuracy and efficiency.

\begin{table}[!h]
\begin{tabular}{c|c|c|c}
\hline  Lightweight/Heavy Classifier & Acc & mAP & Gflops  \\ \hline
ShuffleNetV2\_0.25x Only  & 41.7 & 55.1    &     8.4   \\ 
ShuffleNetV2\_1.0x Only&  62.5  & 73.4 &   23.8    \\ 
ShuffleNetV2\_1.5x Only&  67.2  &  76.8  &   43.0   \\ 
ShuffleNetV2\_2x Only& 70.3    &  79.3    &    72.1     \\ 
ShuffleNetV1\_2x Only &  71.4 &  80.2 &    78.7   \\ \hline
ShuffleNetV2\_0.25x/R(2+1)\_50   & 78.5 & 85.9  &  78.4    \\ 
ShuffleNetV2\_1.0x/R(2+1)\_50   & 79.8 &  86.5 & 100.8     \\ 
ShuffleNetV2\_1.5x/R(2+1)\_50   &  80.2 & 87.1  &  113.0     \\
ShuffleNetV2\_2x/R(2+1)\_50   &  81.0 & 87.6  &   133.2    \\ 
ShuffleNetV1\_2x/R(2+1)\_50   &  \underline{81.2} & \underline{87.7} &  \underline{141.6}   \\ \hline
R(2+1)D\_50 Only &   \textbf{82.2}  &  \textbf{88.5}    &    \textbf{1435.2}     \\ \hline
\end{tabular}
\vspace{0.1cm}
\caption{Results of "a lightweight model only", our proposed Slim-Scan Framework from ``the combinations a lightweight model with a heavy model" and "a heavy model only" framework on ActivityNet-v1.3.}
\label{fig:slim-scan} 

\end{table}

\subsection{Comparison with the State-of-the-art}
\label{sec:experiment3}

\paragraph{\textbf{Comparison with both Clip-Level sampling and Frame-Level sampling frameworks}}

\begin{figure}[!h]
\centering 
\includegraphics[width = 8cm]{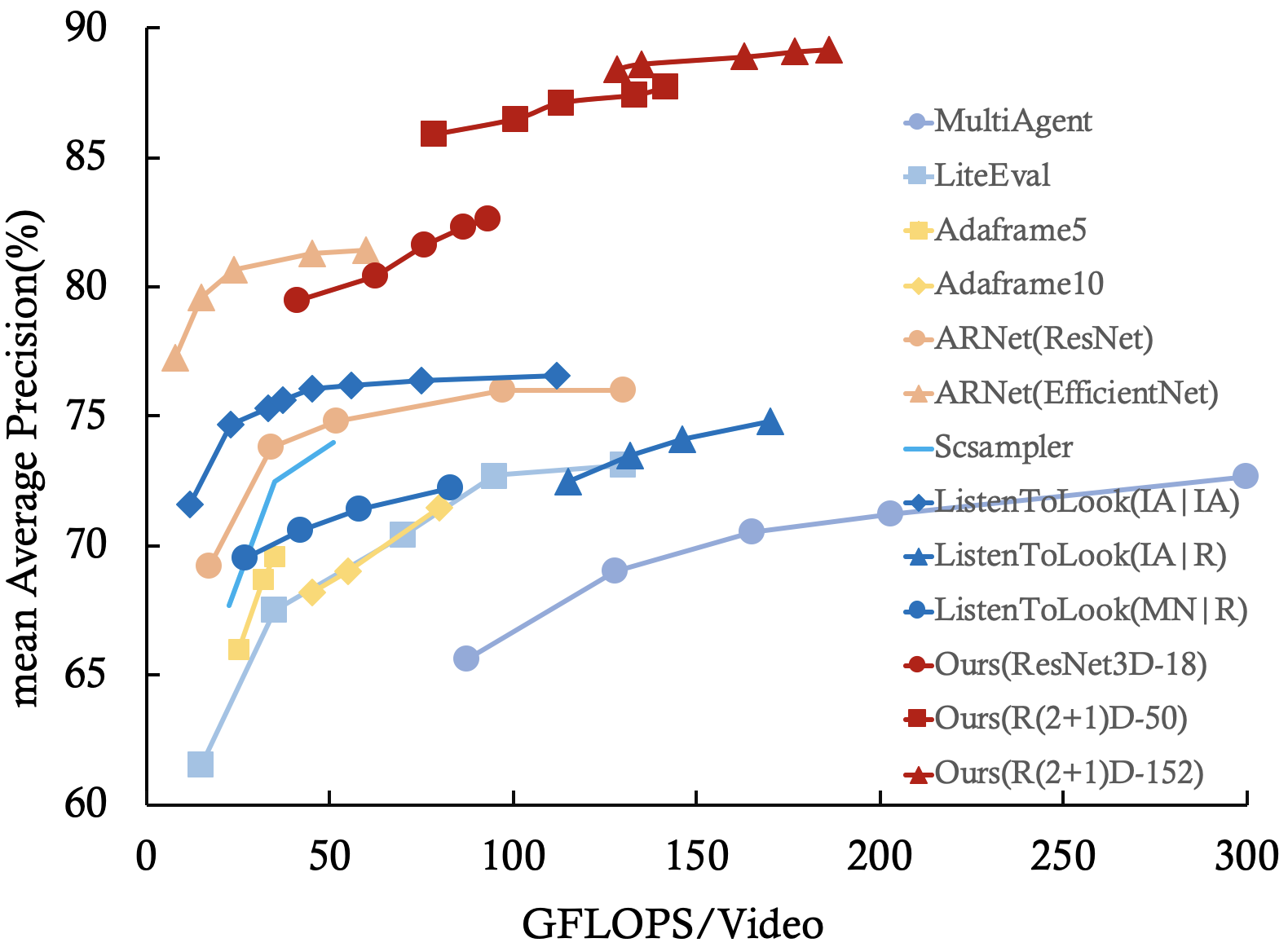}

\caption{Comparison with existing clip-level and frame-level sampling approaches on ActivityNet-v1.3.}
\label{fig:sota}

\end{figure}

Following previous works~\cite{wu2019adaframe,korbar2019scsampler,gao2020listen,meng2020ar} to compare original settings of their frameworks with other methods, we also compare our approach with existing clip-level~\cite{wu2019adaframe, wu2019liteeval, wu2019multi,meng2020ar} and frame-level~\cite{korbar2019scsampler,gao2020listen} methods in Figure~\ref{fig:sota}. The results are reported from their works. To explore the effectiveness of the proposed Slim-Scan framework, we take three kinds of heavy classifiers: R(2+1)D-152, R(2+1)D-50, and ResNet3D-18, combined with five lightweight classifiers mention in Table~\ref{fig:slim-scan}. First, we surpass a large margin~(nearly $10\%$ mAP) compared to other works with similar Gflops (e.g., $100$ Gflops). Our Slim-Scan framework uses 3D CNNs as the backbone, leading to a high calculation cost compared with the 2D CNNs, since the backbone occupied the main part of computation costs. Compared with ARNet~(EfficientNet-2D), we show a slight mAP loss when the computation cost is around $50$ Gflops.  

Also, from the three red lines in Figure~\ref{fig:sota}, we can observe that the heavy classifier determines the upper bound of the mAP. Besides, stronger lightweight models that bring the results closer to the upper bound. In summary, our framework can improve a large margin in mAP with affordable costs, indicating the effectiveness of the Slim-Scan framework.


\paragraph{\textbf{Comparison with the unsampling framework}}
We compare our Skim-Scan framework and Slim-Scan framework with the state-of-the-art work, named \emph{Listen\_to\_Look~(unsampling)}\cite{gao2020listen}. It's an unsampling approach, aggregating the weighted feature based on the attention scores in each timestamp of all clips and getting the final clip-level prediction. From Table~\ref{fig:comparison of attention}, our frameworks can achieve the competitive performance only needs \textbf{$3.2\%$} computational overhead. Besides, they cannot utilize the efficient way of combining lightweight and heavy networks, which leads to the high costs of their overall framework. 

\begin{table}[!h]
\begin{tabular}{c|c|c|c} \hline  
Approaches & Acc & mAP & S\_Gflops/V  \\ \hline
\emph{Listen\_to\_Look~(unsampling)}  & 85.2 & 89.9 & 0.37 \\
Ours~(Skim-Scan)  & \textbf{86.4} & \textbf{90.7} & \textbf{0.012} \\ \hline
\end{tabular}
\vspace{0.1cm}
\caption{Comparison with \emph{Listen\_to\_Look~(unsampling)} in ActivityNet1.3}
\label{fig:comparison of attention}

\end{table}

\subsection{Qualitative Analysis}

\begin{figure}[!h]
\centering 
\includegraphics[width = 7.6cm]{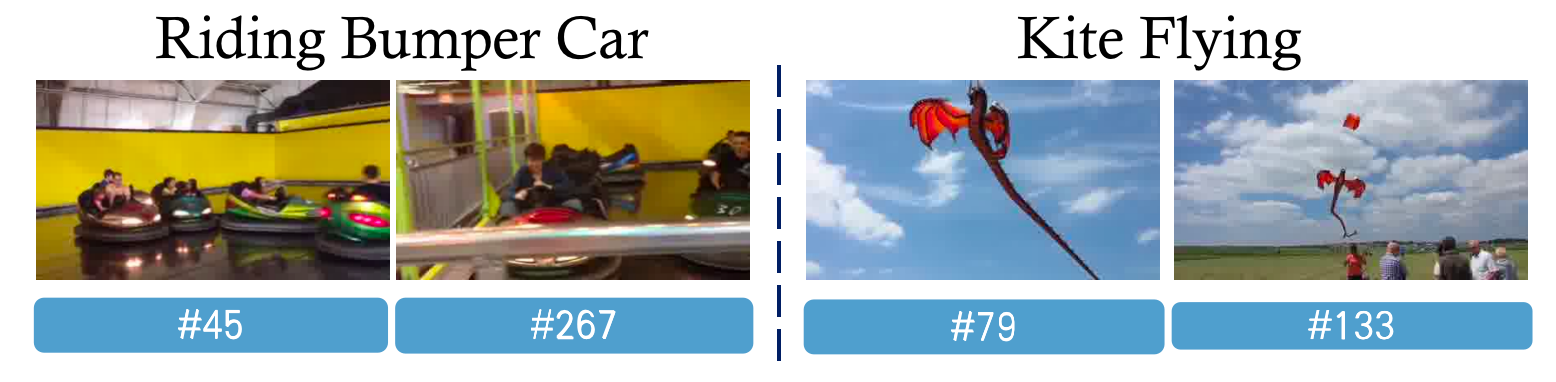}
\caption{Qualitative examples of the adaptive selected clips, illustrating the first frames of the clips, by our method}
\label{fig:case_sample}
\end{figure}

\paragraph{\textbf{Adaptive selection ability}}
In untrimmed videos, the amount of information varies from video to video. Thus, the representative clips should in unequal lengths.
We statistically analyze the number of clips selected in different classes. The class ``Ping Pong'' requires the least number of clips, only $3.75$ clips per video on average, while the class ``Making a cake'' requires the most clips, which needs $42$ clips because of many diverse but misleading clips. Figure ~\ref{fig:case_sample} demonstrates the clips selected by our method. Considering the limited content in these videos, we only select two clips to decide the final predicted label instead of a fixed number of $10$ clips, showing the advantage of our adaptive selecting.


\paragraph{\textbf{Analysis of per-class performance}}
 We statistically analyze the per-class performance to see which classes can benefit more from our strategy compared with \emph{Scsampler} and \emph{Listen to Look}. Specifically, the top $5$ classes in accuracy gain compared with \emph{Scsampler} are: ``Making a cake~($32\%$)'', ``Playing beach volleyball~($25\%$)'', ``Assembling bicycle~($23\%$)'', ``Skiing~($19\%$)'' and ``Braiding hair~($17\%$)''. Compared to the \emph{Listen to Look}, we gain improvements on ``Smoking a cigarette~($36\%$)'', ``Painting~($32\%$)'', ``Shuffleboard~($16\%$)'', ``Hand car wash~($14\%$)'' and ``Drinking coffee~($14\%$)''. We take a deep look into these videos and observe that they all meet interference from misleading and redundant clips, which can be relieved in our framework.
\paragraph{\textbf{Analysis of the failure cases}}
After taking a closer look at failure cases from our methods, we find they are mainly caused by two limitations. 
\begin{enumerate}
    \item \textbf{The limited accuracy of the clip-level prediction.} Because the video-level prediction relies on the quality of clip-level prediction, when a large amount of the clip-level prediction is not consistent with the ground truth label of the video, it is almost impossible to predict the entire video correctly. 
    \item \textbf{Weakness in reasoning.} 
    Our method may predict wrong when the video contains many interference clips. Although our selection framework can cover most contents of videos and consider certain relations among clips, it is hard to reason their primary and secondary relationships between similar inter-class videos. 
\end{enumerate}

Due to limited space in the main paper, we put more detailed qualitative analysis in \emph{supp}.


\section{Conclusion}
This paper proposes an efficient Slim-Scan framework for untrimmed VAR task, containing two parts. First, the Skim-Scan sampling framework guided by information theory, which is model-agnostic and can apply to various backbones on different datasets. It provides us some insights: comprehensive consideration in the content and diversity of information are critical for video understanding. Meanwhile, we propose the class discriminator to drop some misleading clips to boost performance and further reduce the selected clips. Then, we combine a lightweight and heavy models to balance accuracy and efficiency. Comprehensive experiments and analyses demonstrate that our approach using limited clips (30\% reduction) to achieves state-of-the-art results, and also bring 1.0\% $\sim$ 2.9\% improvement for untrimmed video action recognition. We also discuss the limitations and hope this work would inspire more efficient strategies and unsupervised training schemes.

{\small
\bibliographystyle{ieee_fullname}

\begin{thebibliography}{10}\itemsep=-1pt

\bibitem{caba2015activitynet}
Fabian Caba~Heilbron, Victor Escorcia, Bernard Ghanem, and Juan Carlos~Niebles.
\newblock Activitynet: A large-scale video benchmark for human activity
  understanding.
\newblock In {\em Proceedings of the ieee conference on computer vision and
  pattern recognition}, pages 961--970, 2015.

\bibitem{fan2018watching}
Hehe Fan, Zhongwen Xu, Linchao Zhu, Chenggang Yan, Jianjun Ge, and Yi Yang.
\newblock Watching a small portion could be as good as watching all: Towards
  efficient video classification.
\newblock In {\em IJCAI International Joint Conference on Artificial
  Intelligence}, 2018.

\bibitem{x3d2020}
Christoph Feichtenhofer.
\newblock {X3D}: Progressive network expansion for efficient video recognition.
\newblock In {\em {CVPR}}, 2020.

\bibitem{feichtenhofer2019slowfast}
Christoph Feichtenhofer, Haoqi Fan, Jitendra Malik, and Kaiming He.
\newblock Slowfast networks for video recognition.
\newblock In {\em Proceedings of the IEEE international conference on computer
  vision}, pages 6202--6211, 2019.

\bibitem{gao2020listen}
Ruohan Gao, Tae-Hyun Oh, Kristen Grauman, and Lorenzo Torresani.
\newblock Listen to look: Action recognition by previewing audio.
\newblock In {\em Proceedings of the IEEE/CVF Conference on Computer Vision and
  Pattern Recognition}, pages 10457--10467, 2020.

\bibitem{hara2017learning}
Kensho Hara, Hirokatsu Kataoka, and Yutaka Satoh.
\newblock Learning spatio-temporal features with 3d residual networks for
  action recognition.
\newblock In {\em Proceedings of the IEEE International Conference on Computer
  Vision Workshops}, pages 3154--3160, 2017.

\bibitem{hara2018can}
Kensho Hara, Hirokatsu Kataoka, and Yutaka Satoh.
\newblock Can spatiotemporal 3d cnns retrace the history of 2d cnns and
  imagenet?
\newblock In {\em Proceedings of the IEEE conference on Computer Vision and
  Pattern Recognition}, pages 6546--6555, 2018.

\bibitem{he2015deep}
Kaiming He, Xiangyu Zhang, Shaoqing Ren, and Jian Sun.
\newblock Deep residual learning for image recognition. corr abs/1512.03385
  (2015), 2015.

\bibitem{hinton2015distilling}
Geoffrey Hinton, Oriol Vinyals, and Jeff Dean.
\newblock Distilling the knowledge in a neural network.
\newblock {\em arXiv preprint arXiv:1503.02531}, 2015.

\bibitem{hsu2017unsupervised}
Wei-Ning Hsu, Yu Zhang, and James Glass.
\newblock Unsupervised learning of disentangled and interpretable
  representations from sequential data.
\newblock {\em arXiv preprint arXiv:1709.07902}, 2017.

\bibitem{ji20123d}
Shuiwang Ji, Wei Xu, Ming Yang, and Kai Yu.
\newblock 3d convolutional neural networks for human action recognition.
\newblock {\em IEEE transactions on pattern analysis and machine intelligence},
  35(1):221--231, 2012.

\bibitem{jiang2017exploiting}
Yu-Gang Jiang, Zuxuan Wu, Jun Wang, Xiangyang Xue, and Shih-Fu Chang.
\newblock Exploiting feature and class relationships in video categorization
  with regularized deep neural networks.
\newblock {\em IEEE transactions on pattern analysis and machine intelligence},
  40(2):352--364, 2017.

\bibitem{karpathy2014large}
Andrej Karpathy, George Toderici, Sanketh Shetty, Thomas Leung, Rahul
  Sukthankar, and Li Fei-Fei.
\newblock Large-scale video classification with convolutional neural networks.
\newblock In {\em Proceedings of the IEEE conference on Computer Vision and
  Pattern Recognition}, pages 1725--1732, 2014.

\bibitem{kataoka2020would}
Hirokatsu Kataoka, Tenga Wakamiya, Kensho Hara, and Yutaka Satoh.
\newblock Would mega-scale datasets further enhance spatiotemporal 3d cnns?
\newblock {\em arXiv preprint arXiv:2004.04968}, 2020.

\bibitem{kopuklu2019resource}
Okan K{\"o}p{\"u}kl{\"u}, Neslihan Kose, Ahmet Gunduz, and Gerhard Rigoll.
\newblock Resource efficient 3d convolutional neural networks.
\newblock In {\em 2019 IEEE/CVF International Conference on Computer Vision
  Workshop (ICCVW)}, pages 1910--1919. IEEE, 2019.

\bibitem{korbar2019scsampler}
Bruno Korbar, Du Tran, and Lorenzo Torresani.
\newblock Scsampler: Sampling salient clips from video for efficient action
  recognition.
\newblock In {\em Proceedings of the IEEE International Conference on Computer
  Vision}, pages 6232--6242, 2019.

\bibitem{kullback1951information}
Solomon Kullback and Richard~A Leibler.
\newblock On information and sufficiency.
\newblock {\em The annals of mathematical statistics}, 22(1):79--86, 1951.

\bibitem{lea2017temporal}
Colin Lea, Michael~D Flynn, Rene Vidal, Austin Reiter, and Gregory~D Hager.
\newblock Temporal convolutional networks for action segmentation and
  detection.
\newblock In {\em proceedings of the IEEE Conference on Computer Vision and
  Pattern Recognition}, pages 156--165, 2017.

\bibitem{locatello2020commentary}
Francesco Locatello, Stefan Bauer, Mario Lucic, Gunnar R{\"a}tsch, Sylvain
  Gelly, Bernhard Sch{\"o}lkopf, and Olivier Bachem.
\newblock A commentary on the unsupervised learning of disentangled
  representations.
\newblock In {\em Proceedings of the AAAI Conference on Artificial
  Intelligence}, volume~34, pages 13681--13684, 2020.

\bibitem{maxwell1972skimming}
Martha~J Maxwell.
\newblock Skimming and scanning improvement: the needs, assumptions and
  knowledge base.
\newblock {\em Journal of Reading Behavior}, 5(1):47--59, 1972.

\bibitem{meng2020ar}
Yue Meng, Chung-Ching Lin, Rameswar Panda, Prasanna Sattigeri, Leonid
  Karlinsky, Aude Oliva, Kate Saenko, and Rogerio Feris.
\newblock Ar-net: Adaptive frame resolution for efficient action recognition.
\newblock {\em arXiv preprint arXiv:2007.15796}, 2020.

\bibitem{puiutta2020explainable}
Erika Puiutta and Eric Veith.
\newblock Explainable reinforcement learning: A survey.
\newblock {\em arXiv preprint arXiv:2005.06247}, 2020.

\bibitem{qiu2017learning}
Zhaofan Qiu, Ting Yao, and Tao Mei.
\newblock Learning spatio-temporal representation with pseudo-3d residual
  networks.
\newblock In {\em proceedings of the IEEE International Conference on Computer
  Vision}, pages 5533--5541, 2017.

\bibitem{robinson2008entropy}
Derek~W Robinson.
\newblock Entropy and uncertainty.
\newblock {\em Entropy}, 10(4):493--506, 2008.

\bibitem{shannon2001mathematical}
Claude~Elwood Shannon.
\newblock A mathematical theory of communication.
\newblock {\em ACM SIGMOBILE mobile computing and communications review},
  5(1):3--55, 2001.

\bibitem{tan2019efficientnet}
Mingxing Tan and Quoc~V Le.
\newblock Efficientnet: Rethinking model scaling for convolutional neural
  networks.
\newblock {\em arXiv preprint arXiv:1905.11946}, 2019.

\bibitem{tran2015learning}
Du Tran, Lubomir Bourdev, Rob Fergus, Lorenzo Torresani, and Manohar Paluri.
\newblock Learning spatiotemporal features with 3d convolutional networks.
\newblock In {\em Proceedings of the IEEE international conference on computer
  vision}, pages 4489--4497, 2015.

\bibitem{tran2019video}
Du Tran, Heng Wang, Lorenzo Torresani, and Matt Feiszli.
\newblock Video classification with channel-separated convolutional networks.
\newblock In {\em Proceedings of the IEEE International Conference on Computer
  Vision}, pages 5552--5561, 2019.

\bibitem{tran2018closer}
Du Tran, Heng Wang, Lorenzo Torresani, Jamie Ray, Yann LeCun, and Manohar
  Paluri.
\newblock A closer look at spatiotemporal convolutions for action recognition.
\newblock In {\em Proceedings of the IEEE conference on Computer Vision and
  Pattern Recognition}, pages 6450--6459, 2018.

\bibitem{wu2019multi}
Wenhao Wu, Dongliang He, Xiao Tan, Shifeng Chen, and Shilei Wen.
\newblock Multi-agent reinforcement learning based frame sampling for effective
  untrimmed video recognition.
\newblock In {\em Proceedings of the IEEE International Conference on Computer
  Vision}, pages 6222--6231, 2019.

\bibitem{wu2019liteeval}
Zuxuan Wu, Caiming Xiong, Yu-Gang Jiang, and Larry~S Davis.
\newblock Liteeval: A coarse-to-fine framework for resource efficient video
  recognition.
\newblock In {\em Advances in Neural Information Processing Systems}, pages
  7780--7789, 2019.

\bibitem{wu2019adaframe}
Zuxuan Wu, Caiming Xiong, Chih-Yao Ma, Richard Socher, and Larry~S Davis.
\newblock Adaframe: Adaptive frame selection for fast video recognition.
\newblock In {\em Proceedings of the IEEE Conference on Computer Vision and
  Pattern Recognition}, pages 1278--1287, 2019.

\bibitem{split2020}
Ailing Zeng, Xiao Sun, Fuyang Huang, Minhao Liu, Qiang Xu, and Stephen Lin.
\newblock Srnet: Improving generalization in 3d human pose estimation with a
  split-and-recombine approach.
\newblock In {\em European Conference on Computer Vision}, pages 507--523.
  Springer, 2020.

\bibitem{zolfaghari2018eco}
Mohammadreza Zolfaghari, Kamaljeet Singh, and Thomas Brox.
\newblock Eco: Efficient convolutional network for online video understanding.
\newblock In {\em Proceedings of the European conference on computer vision
  (ECCV)}, pages 695--712, 2018.

\end{thebibliography}
}


\pagebreak

\twocolumn[{
\renewcommand\twocolumn[1][]{#1}
\begin{center}
\textbf{\Large Supplementary Material:\\Skimming and Scanning for Untrimmed Video Action Recognition}\end{center}
}]
\vspace{10pt}
\setcounter{equation}{0}
\setcounter{figure}{0}
\setcounter{table}{0}
\setcounter{section}{0}
\setcounter{page}{0}
\maketitle

In this supplementary material, we first present more quantitative results and analyses to show the effectiveness of our Skim-Scan Framework in Sec.~\ref{sec:results}. Then, in Sec.~\ref{sec:design}, we discuss more specific design in our framework. Next, we further discuss the limitations in this work in Sec.~\ref{sec:fail}. Finally, based on the above discussion, we give some future work in Sec.~\ref{sec:future}. All results are based on the backbone R(2+1)D-152 in ActivityNet-v1.3 dataset. \emph{All the visualization examples are shown at the end of this manuscript due to their large size.} 

\section{Additional Result and Analysis}
\label{sec:results}
In this section, we analyze the impact of adaptive selection and the per-class performance of our framework in Sec.~\ref{sec:ada} and Sec.~\ref{sec:per}, respectively. 

\subsection{The Ability of Adaptive Selection}
\label{sec:ada}
\begin{figure}[!h]
\centering 
\includegraphics[width = 8cm]{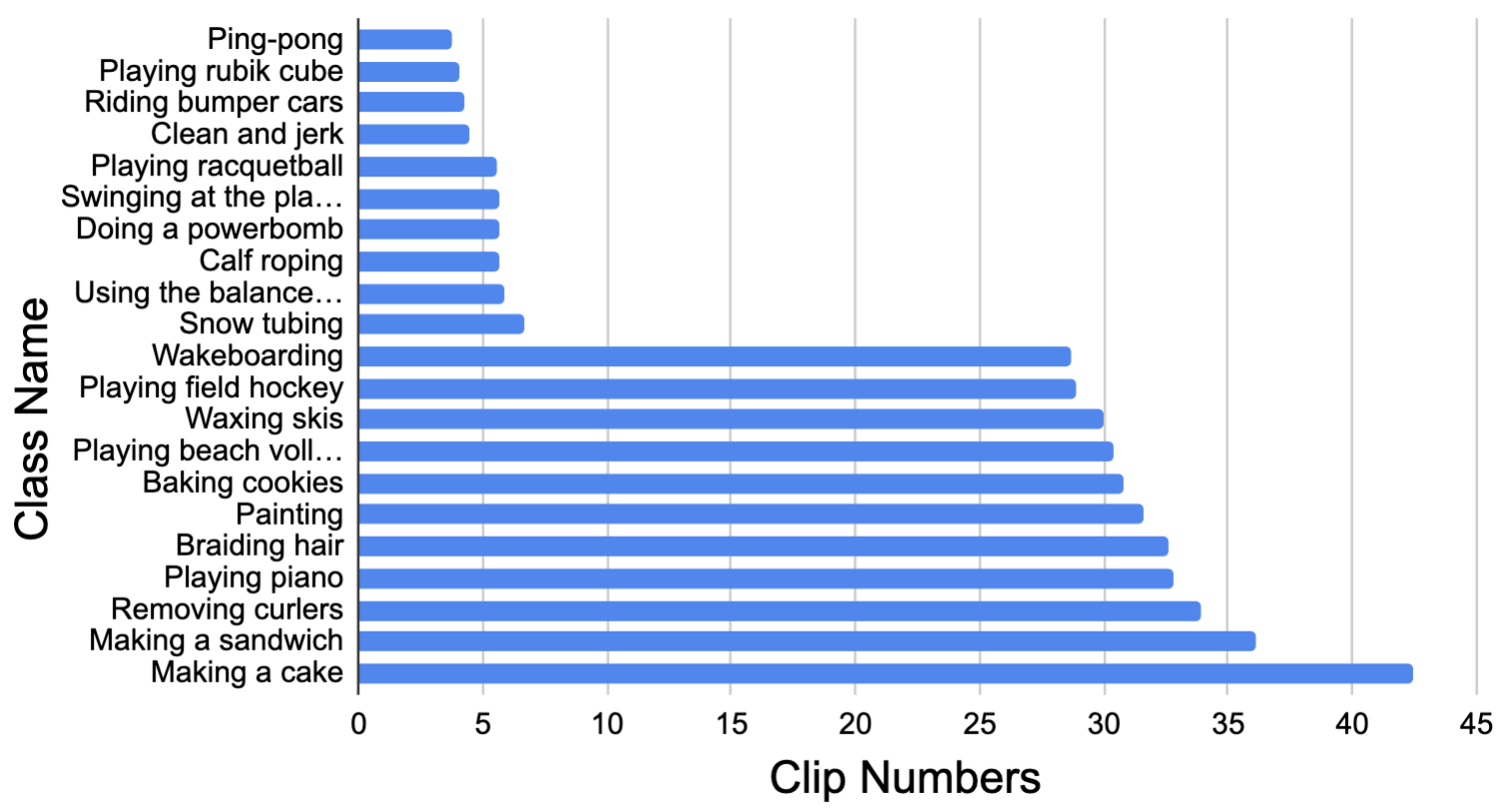}
\caption{The Top-10 classes that require the most and the least number of clips in average.}
\label{fig:num_class}
\end{figure}

Compared with previous works~\cite{gao2020listen,korbar2019scsampler}, which select a fixed number of clips, our approach allows for more flexible clip selection. Accordingly, we statistically analyze the number of clips selected in different categories. Figure~\ref{fig:num_class} shows the Top-10 classes that require the most and the least number of clips. The number of clips required by different classes of videos varies significantly, which is affected by not only the content of each intra-class video but also the interference from inter-class videos. 

\begin{figure*}[!h]
\centering 
\includegraphics[width = 14.6cm]{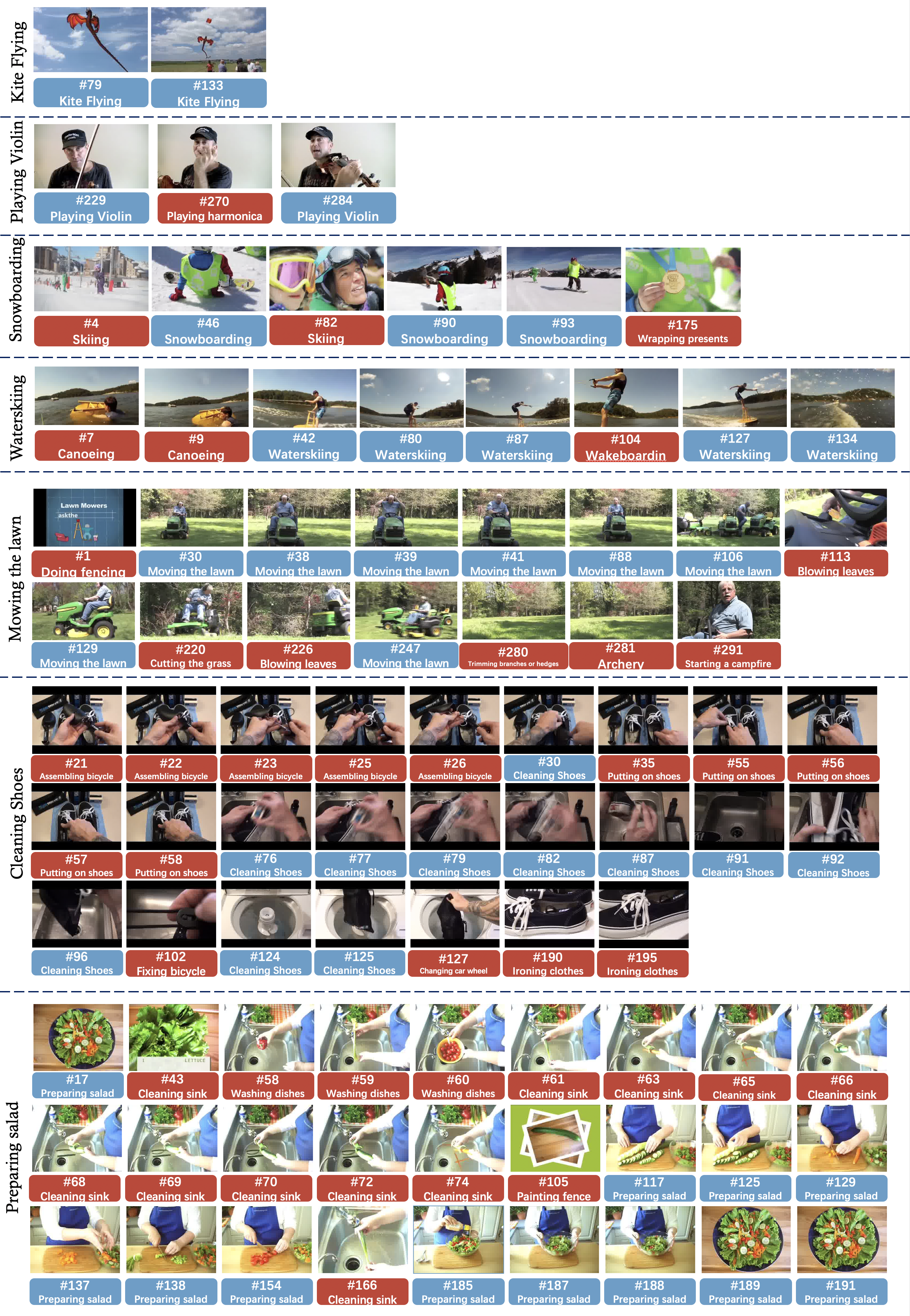}
\caption{Qualitative results of video samples with various number of selected clips. We utilize the first image of each clip to represent the information of the entire clip, and indicate the index of the corresponding clip and its predicted label at the same time. The blue represents the positive clips, and the red represents the negative clips. The corresponding labels for those videos are in the leftmost vertical row.}
\label{fig:num_class_vis}
\end{figure*}

We visualize some cases in Figure~\ref{fig:num_class_vis} and have the following two observations. On the one hand, when the video content is uniform and clear, the proposed selection strategy only needs a few clips to cover the complete information and make reasonable predictions. For instance, in the first two rows of Figure~\ref{fig:num_class_vis}, only 2-3 clips are needed to represent the entire video. On the other hand, when our method is faced with multiple scene-switching and interference situations, shown in the last two cases in Figure~\ref{fig:num_class_vis}, it will select more clips and uses these clips to make a final prediction. 

\subsection{Per-Class Performance Analysis}
\label{sec:per}

Accordingly, we compare the per-class performance of our Skim-Scan framework with previous works~\cite{gao2020listen,korbar2019scsampler}, wherein we show the Top-10 classes with the most accuracy gap (positive and negative) between our solution and the other works. In this subsection, we mainly analyze these videos that our solution performs better than others, and we will discuss the worse cases in Sec.~\ref{sec:fail} as our failure cases.

\begin{figure}[!h]
\centering 
\includegraphics[width = 8cm]{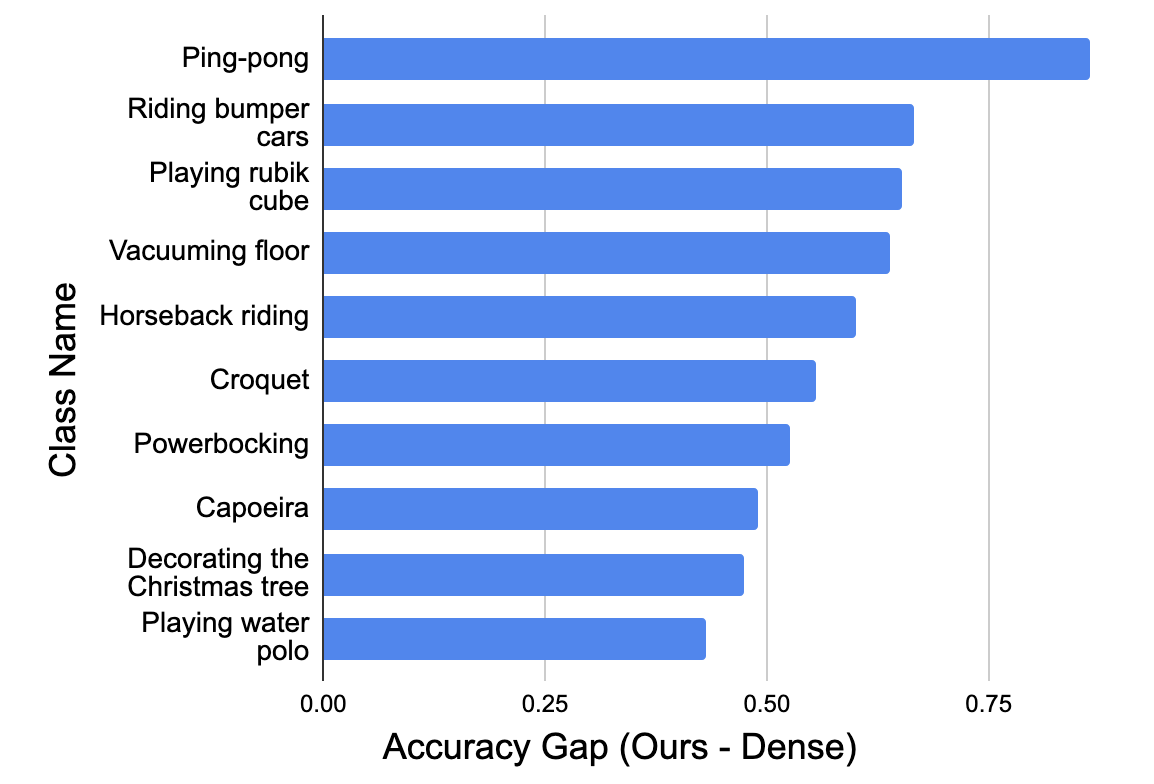}
\caption{The Top-10 classes with the most difference of the Accuracy (positive and negative) between Ours and \emph{Dense}}.
\label{fig:acc_dense}
\end{figure}

\begin{figure*}[!h]
\centering 
\includegraphics[width = 16cm]{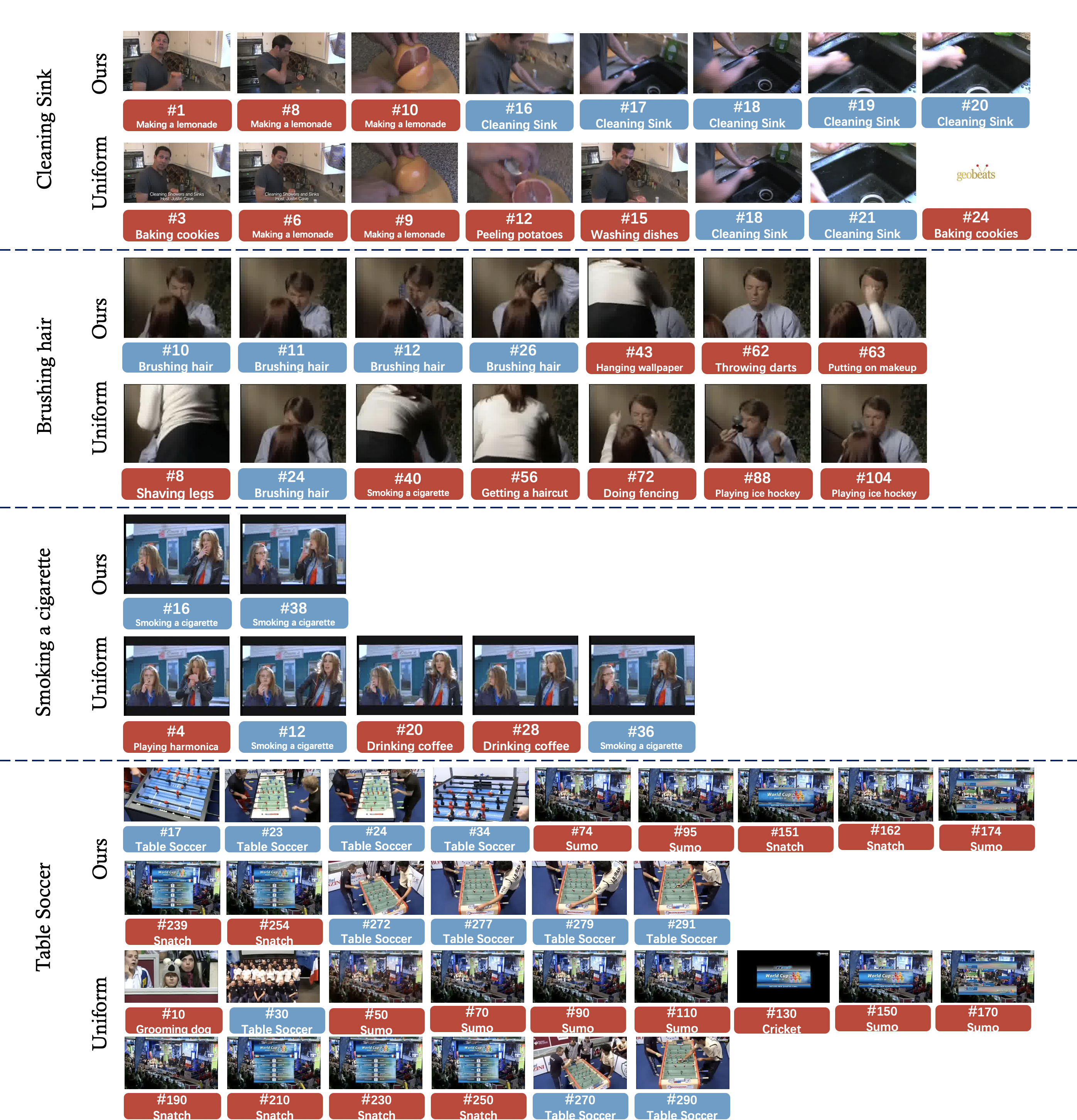}
\caption{Qualitative results of video samples which is predicted correctly by Ours but failed by \emph{Dense}. Due to space constraints, we do not display all the clips in the video, we choose the same clip number of our method to uniform select from the video to roughly represent the information of the video. We utilize the first image of each clip to represent the information of the entire clip, and indicate the index of the corresponding clip and its predicted label at the same time. The blue represents the positive clips, and the red represents the negative clips. The corresponding labels for those videos are in the leftmost vertical row.}
\label{fig:dense_vis}
\end{figure*}

\paragraph{\textbf{Per-Class Performance compared to \emph{Dense}}}
\emph{Dense} strategy averages the clip-level prediction scores from all clips with equal consideration to get the video-level prediction.
Figure~\ref{fig:acc_dense} shows the comparison results. By analyzing those videos where our results are significantly better than \emph{Dense}, we find that there are $78\%$ clips with entropy higher than our entropy threshold in these videos on average, while the average ratio for all the videos is only $42\%$. Moreover, we can observe that most high-entropy-score clips are negative clips, which harm the final prediction. However, our proposed \emph{skimming with entropy elimination} can help in the final prediction of these videos by reducing the number of clips required and significantly suppressing some of the uninformative and irrelevant clips with high entropy scores, showing the significance of the skimming strategy.
More qualitative analysis for the video samples is presented in Figure~\ref{fig:dense_vis}.




\begin{figure}[!h]
\centering 
\includegraphics[width = 8cm]{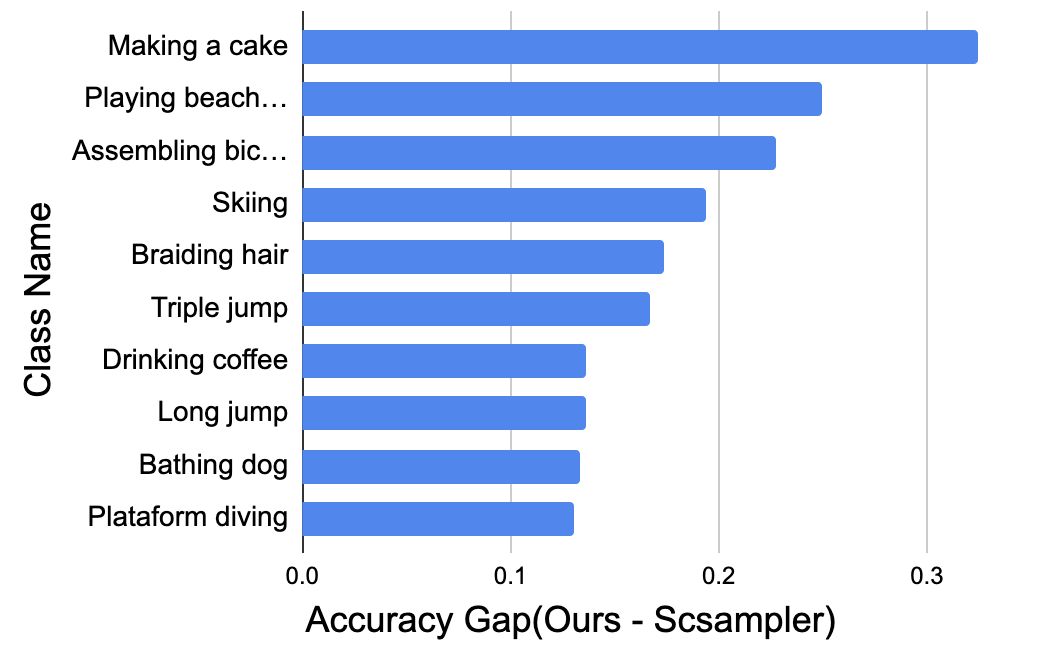}
\caption{The Top-10 classes with the most difference of the Accuracy (positive and negative) between Ours and \emph{Scsampler}}.
\label{fig:acc_scsampler}
\end{figure}


\paragraph{\textbf{Per-Class Performance compared to \emph{Scsampler}}}
\emph{Scsampler}~\cite{korbar2019scsampler} proposes a deep neural network to determine the ``saliency score'' of all clips and selects the $N$ most salient clips from all clips in each video. $N$ is usually fixed as 10.
We show the comparison results in Figure~\ref{fig:acc_scsampler}.

\begin{figure*}[!h]
\centering 
\includegraphics[width = 16cm]{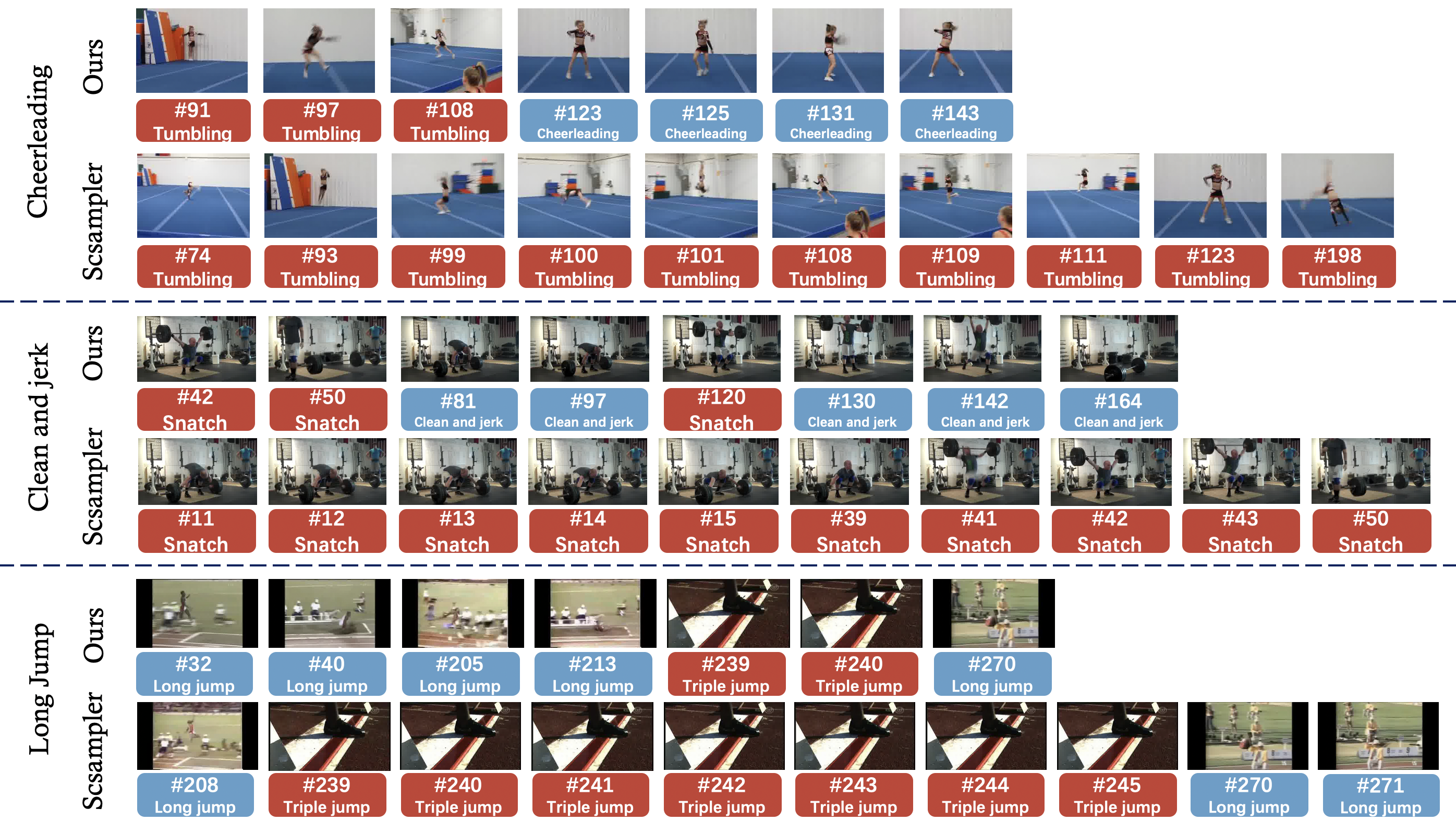}
\caption{Qualitative results of video samples which is predicted correctly by Ours but failed by \emph{Scsampler}. We utilize the first image of each clip to represent the information of the entire clip, and indicate the index of the corresponding clip and its predicted label at the same time. The blue represents the positive clips, and the red represents the negative clips. The corresponding labels for those videos are in the leftmost vertical row.}
\label{fig:scsampler_vis}
\end{figure*}

By analyzing the videos predicted successfully by our framework but failed by \emph{Scsampler}, we find that these videos often contain diverse information and have interference clips belonging to different classes, like ``Making a cake". Moreover, \emph{Scsampler} only pays attention to the clips with the highest salience scores with ``only look once" selection strategy, which may contain a certain degree of redundancy and ignore the complex relations among clips. Some cases are shown in Figure~\ref{fig:scsampler_vis}. Besides, because of the fixed number of selection limitations, they may miss some more representative clips with relatively lower salience scores, leading to wrong predictions. In contrast, in our Skim-Scan framework, we will first drop uninformative and some misleading clips, and then select diverse clips with JS divergence in a ``divide and conquer" strategy. Thus, we can cover the essential information to make more thoughtful predictions. The comparison between ours and \emph{Scsampler} indicates the superiority of scanning by the proposed JS Divergence and adaptive selection of different videos.

\begin{figure}[!h]
\centering 
\includegraphics[width = 8.5cm]{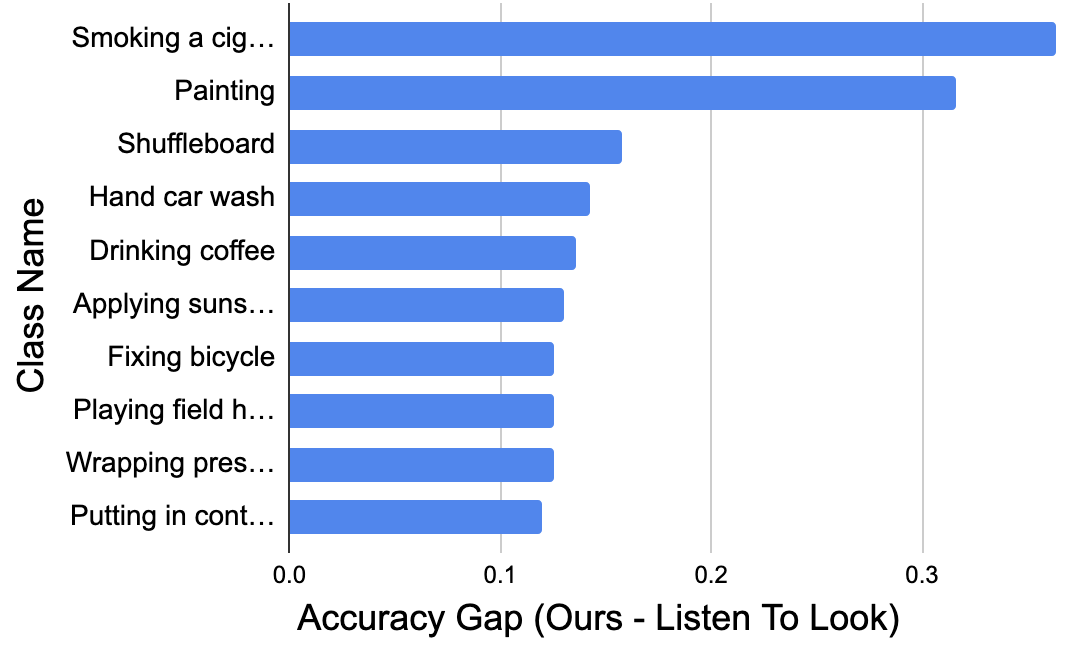}
\caption{The Top-10 classes with the most difference of the Accuracy (positive and negative) between Ours and \emph{Listen to Look~(unsampling)}}.

\label{fig:acc_listen_to_look}
\end{figure}
\begin{figure*}[!h]
\centering 
\includegraphics[width = 16cm]{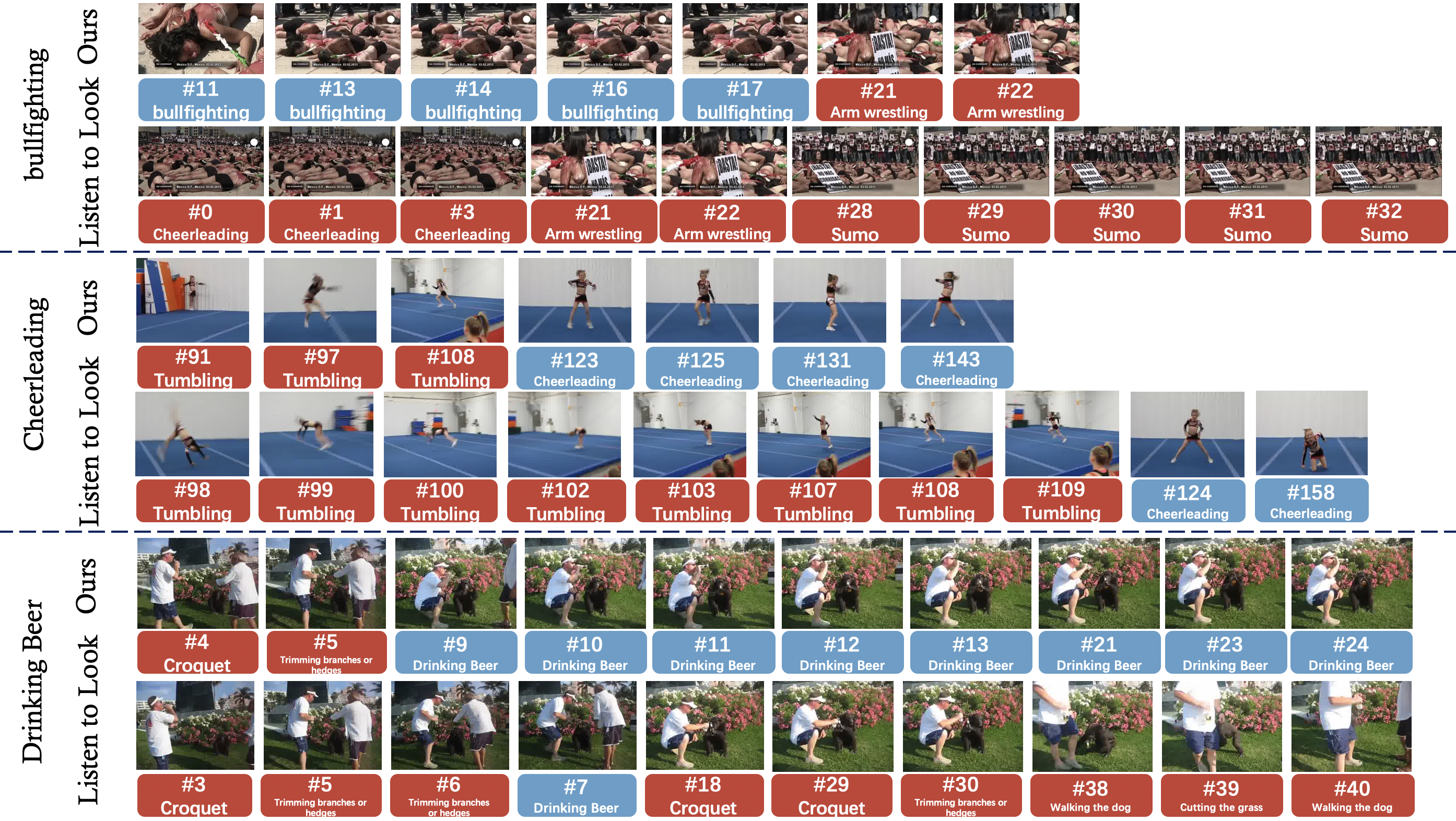}
\caption{Qualitative results of video samples which is predicted correctly by Ours but failed by \emph{Listen to Look}. We utilize the first image of each clip to represent the information of the entire clip, and indicate the index of the corresponding clip and its predicted label at the same time. The blue represents the positive clips, and the red represents the negative clips. The corresponding labels for those videos are in the leftmost vertical row.}
\label{fig:listen_vis}
\end{figure*}

\paragraph{\textbf{Per-Class Performance compared to \emph{Listen to Look~(Unsampling)}}}
\emph{Listen to Look}~\cite{gao2020listen} is the state-of-the-art strategy to aggregate the weighted features by the attention scores in each clip to get the final prediction. The comparison results and the corresponding visualization samples are shown in Figure~\ref{fig:acc_listen_to_look} and Figure~\ref{fig:listen_vis}, respectively.

\begin{figure}[htbp]\centering                              
\subfigure[Clip number of the same class]{                    
\begin{minipage}{4cm}\centering \includegraphics[width=4cm]{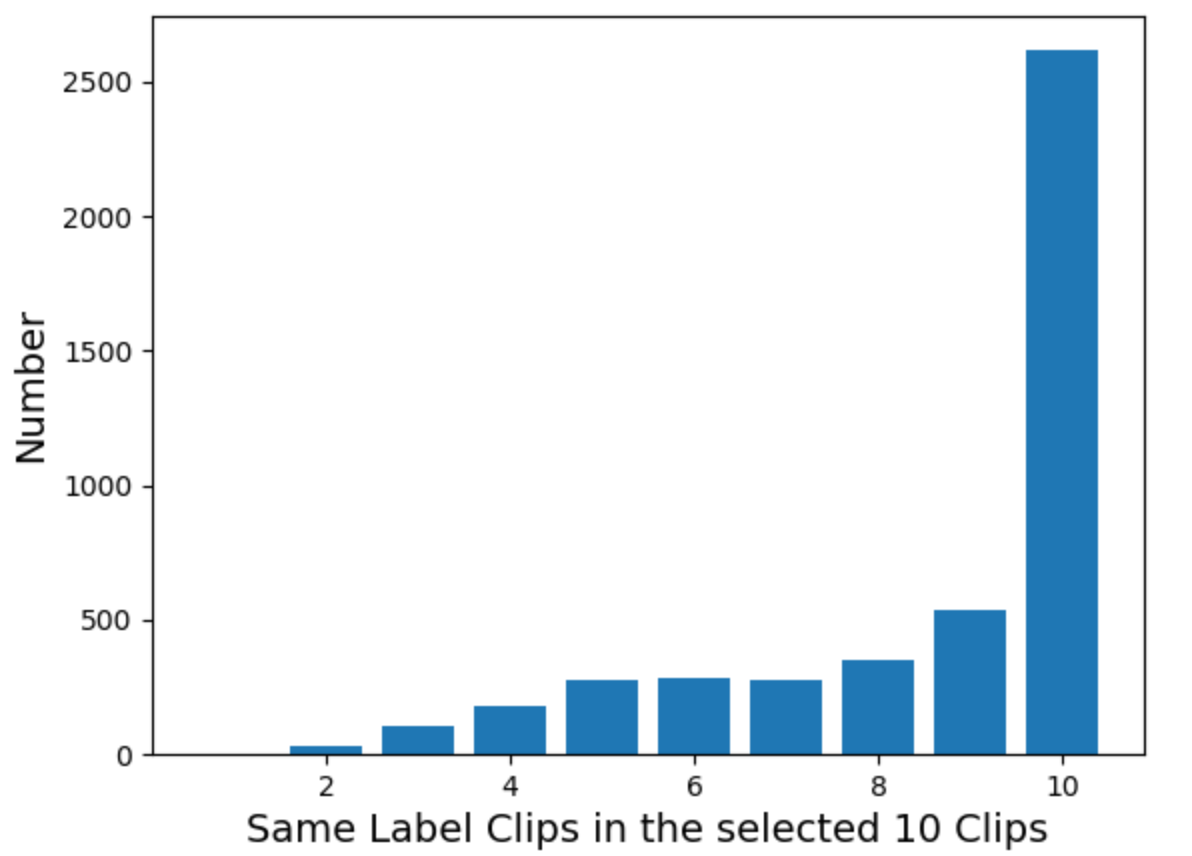}     
\end{minipage}}
\subfigure[Std. of attention score]{
\begin{minipage}{4cm}\centering \includegraphics[width=4cm]{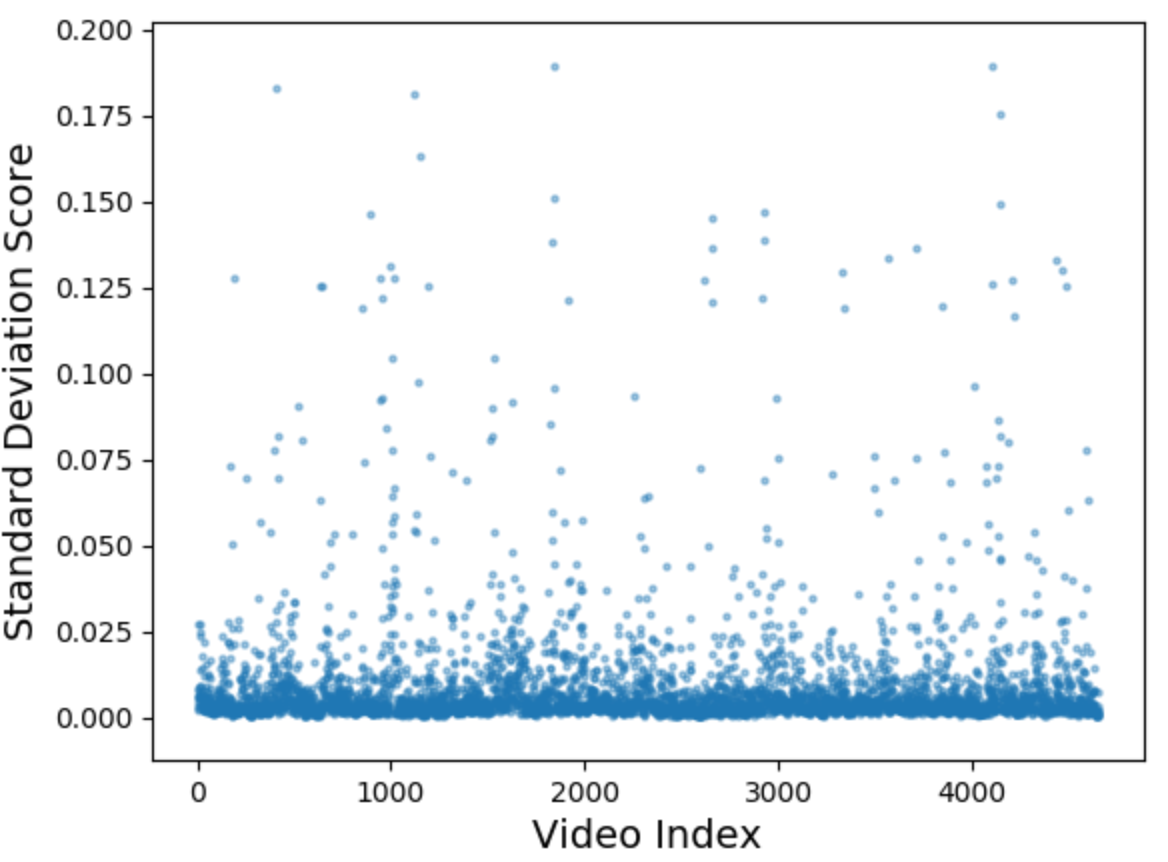}
\end{minipage}}
\caption{(a) shows the statistics of the maximum number of clips with the same clip-level class among the 10 selected clips by \emph{Listen to Look (unsampling)} in each video. (b) shows the scatter plot of standard deviation (Std.) of attention scores of all clips in each videos.}                       
\label{fig:std_value}                                            
\end{figure}

We find that the failure videos from \emph{Listen to Look} are similar to those from \emph{Scsampler}, which often contain diverse information and have misleading clips belonging to different classes. In Figure~\ref{fig:std_value}(a), we calculate he maximum number of clips with the same clip-level class in their method if only the first $10$ clips are selected. We find that their method of calculating the attention scores mainly selects the most similar clips, and relatively few cases could select a variety of clips, leading to some limitations of their method in selecting representative clips. Besides, we take a closer look at the predicted attention scores of clips among the videos in \emph{Listen to Look (Unsampling)}. As shown in Figure~\ref{fig:std_value}, the standard deviation of attention scores in most videos is quite small. In other words, the distribution of the attention scores is quite uniform, which does not give obvious preferences to any clips, undermining its effectiveness. 

\emph{Listen to Look} also has a relatively high computational cost due to their complexity. Compared with it, our Skim-Scan framework is more stable, interpretable, and efficient.


\section{Design Block Analysis}
\label{sec:design}
In this section, we first analyze the three related metrics to approximate the Information Gain in the scanning stage in Sec.~\ref{sec:metics}. Then, we discuss the benefit of the class label information in the proposed class discriminator in Sec.~\ref{sec:binary}.

\subsection{Approximate Measure of Information Gain}
\label{sec:metics}
Several well-known metrics could be used to measure the distance between two distributions, and hence to approximate the Information Gain, i.e., \emph{KL divergence}, \emph{Wasserstein Distance}, and \emph{JS Divergence}. 

Table~\ref{table:metrics_for_IG} shows the performance of our selection strategy with these metrics. First, different metrics give the same accuracy and mAP but require a different number of clips. Second, the calculation costs of \emph{KL Divergence} and \emph{JS Divergence} are similar (JS Divergence needs less clip number than KL Divergence). However, the computational cost of \emph{Wasserstein Distance} is much higher, about 30 times longer than the JS Divergence, with more clip number.

Third, we need to calculate the information gain of the selected clips and the to-be-selected clips, respectively, so that it is possible to select it in an order-independent manner for each clip. 

Considering the above factors, we finally utilize \emph{JS Divergence} to approximate the information gain in our Skim-Scan framework.


\begin{table}[!h]
\centering
\begin{tabular}{c|ccc} 
\hline
  Metrics& Acc. & mAP & \begin{tabular}[c]{@{}c@{}}Clip\\ Num.\end{tabular}  \\ \hline
KL Divergence &86.4 & 90.7 & 10.3      \\  \hline
Wasserstein Distance & 86.4  & 90.7 &  12.4     \\  \hline
JS Divergence (Ours)   & 86.4 & 90.7 &  \textbf{9.2}   \\ \hline
\end{tabular}
\vspace{0.1cm}
\caption{Performance of our Skim-Scan Selection framework with various metrics to approximate Information Gain. Acc. and Num. are the abbreviation for Accuracy and Number, respectively.}
\label{table:metrics_for_IG}
\end{table}

\begin{table}[!h]
\centering
\begin{tabular}{c|ccc} 
\hline
Method  & Acc. & mAP &  \begin{tabular}[c]{@{}c@{}}Clip\\ Num.\end{tabular}   \\ \hline
Without C.D.  &83.5 & 88.9 & 205.3     \\  \hline
Plain C.D.  &  85.0 & 89.8 &  165.4    \\  \hline
Conditional C.D. (Ours) & \textbf{85.2} & \textbf{89.9} & \textbf{143.4} \\  \hline
Oracle C.D. & 85.3 & 90.1 & 135.0 \\  \hline
\end{tabular}
\vspace{0.1cm}
\caption{\textbf{Dense} performance w/wo various Class Discriminator (Abbreviated as C.D.). }
\label{table:Binary_Classifier_of_Dense}
\end{table}

\subsection{Additional Analysis of the Class Discriminator}
\label{sec:binary}
We propose a Class Discriminator to classify the unannotated clips and the annotated clips of each video and then reduce the interference due to unannotated clips. Our class discriminator concatenates a one-hot vector representing the video's class label with the videos' feature as the input. To show the necessity of the class-label information, we compare our method with three different settings in Table~\ref{table:Binary_Classifier_of_Dense}. \emph{Without C.D.} is the setting from \emph{dense} clips without any samplings. \emph{Plain C.D.} means the class discriminator without combining with the class labels of the video. \emph{Oracle C.D.} directly drops all unannotated clips by ground truth binary labels, which indicates the upper bound of this method. 

From Table~\ref{table:Binary_Classifier_of_Dense}, we can observe that \emph{Plain C.D.} achieves significant improvement over \emph{Without C.D.}, justifying the benefits of the binary classifier to drop some misleading clips. Moreover, \emph{Conditional C.D.} gains further $0.2\%$ compared with \emph{Plain C.D.}, proving the effectiveness of the class-label information.

\section{Failure Case Analysis}
\label{sec:fail}
Based on the two limitations shown in the main paper, we further analyze the causes of errors here in statistical results and some visualization results.

\begin{figure*}[!h]
\centering 
\includegraphics[width = 16cm]{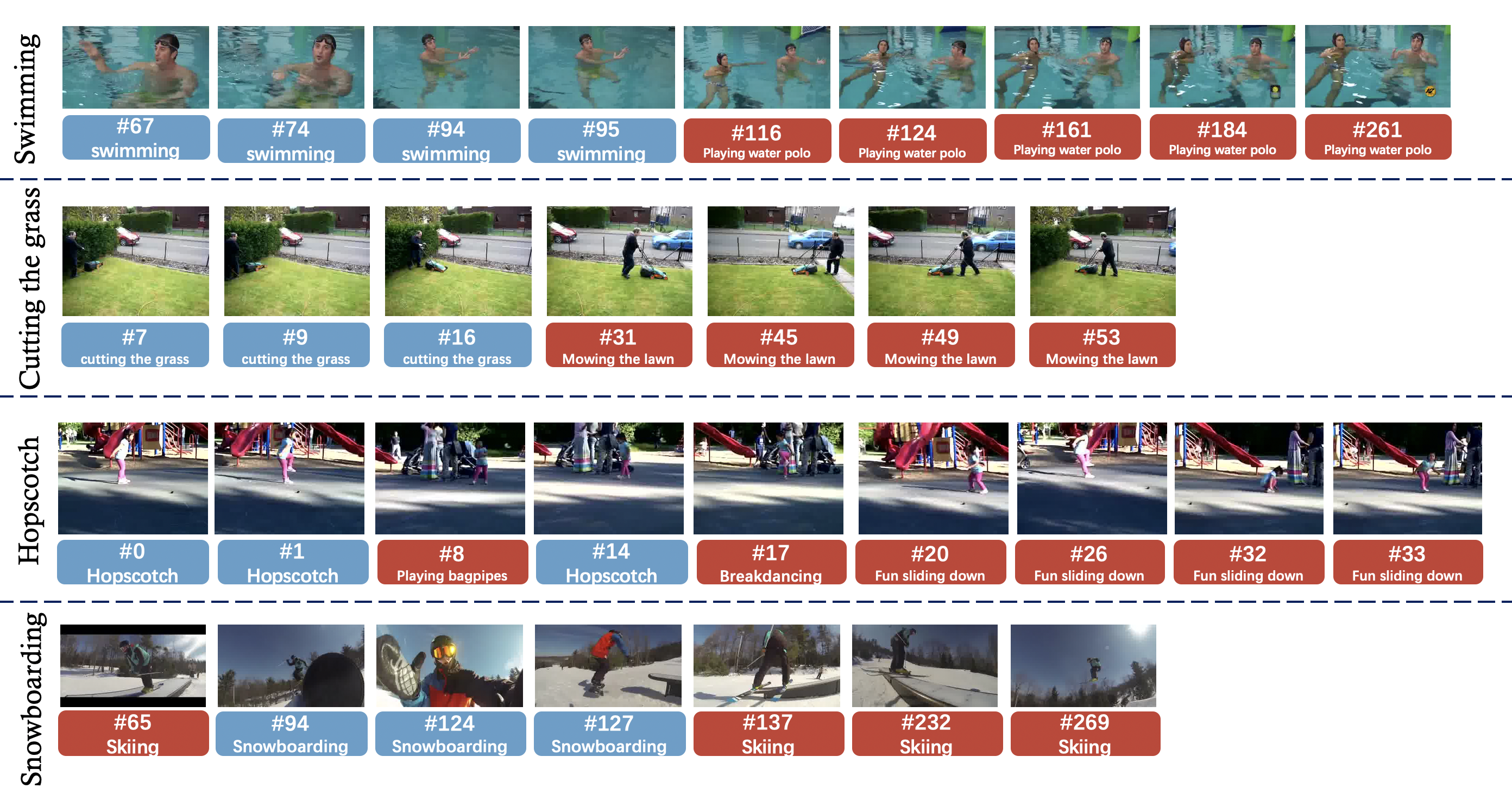}
\caption{Qualitative results of the failure cases in our selection strategy. We utilize the first image of each clip to represent the information of the entire clip, and indicate the index of the corresponding clip and its predicted label at the same time. The blue represents the positive clips, and the red represents the negative clips. The corresponding labels for those videos are in the leftmost vertical row.}
\label{fig:wrong_vis}
\end{figure*}

\begin{enumerate}
    \item \textbf{The limited capability of the clip-level classifier.} In Figure~\ref{fig:percent}, we demonstrate the percentage of the videos predicted correctly under the corresponding portion of positive clips. We can observe \emph{a positive correlation} between the percentage of positive clips and the percentage of correct prediction in the video level from the class classifier. Specifically, when the percentage of positive clips is quite low, the videos are hard to predict well. In summary, the video-level prediction relies on the clip-level prediction accuracy, which should be improved for better video-level performance. Therefore, many video-level prediction failure cases are due to the limited capability of the clip-level classifier. 
    \begin{figure}[!h]
    \centering 
    \includegraphics[width = 8cm]{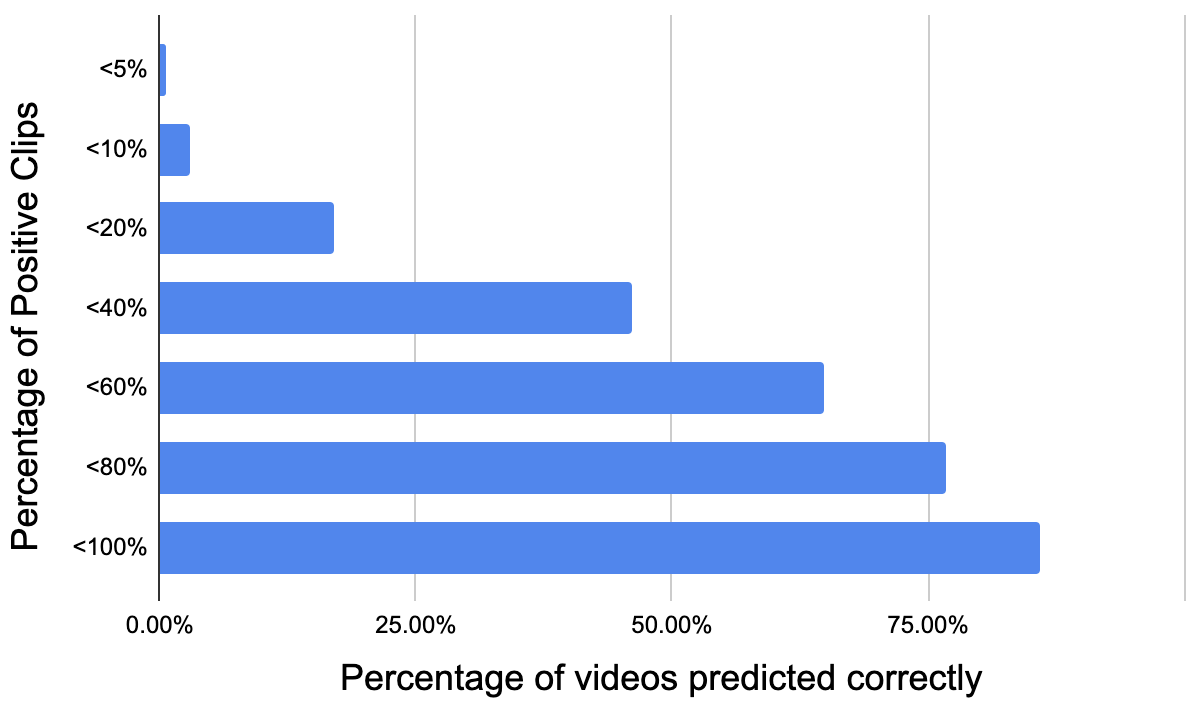}
    \caption{Illustration the correlation between the percentage of positive clips and the percentage of correct. video-level prediction from the backbone R(2+1)D-152.}
    \label{fig:percent}
    \end{figure}

    \item \textbf{Weakness in reasoning.} 
    Although our sampling framework can cover most essential contents of videos and consider certain relations among clips, it is still hard to reason their primary and secondary relationships between similar inter-class videos. The reason here is because when we select these clips by our framework, we directly average the features of these clips to get the final result, without considering then giving some different weights to these features to distinguish their importance in the current video further.
    As shown in the first case of Figure~\ref{fig:wrong_vis}, if we can understand the differences among clips, we could classify it as ``Swimming'' instead of ``Playing water polo''. The partial clips predicted as ``Playing water polo'' is reasonable, but there are no balls and only two people in \emph{all clips}, which is the key to distinguishing the two activities. However, our method only averages the selected clip-level features to obtain the final video-level prediction and shows limitations in effective reasoning. Thus, it makes the wrong video-level predictions.   
    
\end{enumerate}

\section{Future Work}
\label{sec:future}
After analyzing our framework and its failure cases, we give two future directions in response to the above problems.

For one thing, since there are inevitably various complexities in untrimmed videos, we have discussed in this work how to handle some of them \emph{in a disentangled way} and show their effectiveness. In the future, an interesting direction is how to use the ``divide-and-conquer" idea in designing the backbone network to both improve the performance and decrease the computational costs.

For another, understanding the local and global relationships among the intra-class clips better and fuse their features appropriately will be important. 

\end{document}